\documentclass{article}

\usepackage{microtype}
\usepackage{graphicx}
\usepackage{subcaption}
\usepackage{booktabs} 
\usepackage{graphicx} 
\usepackage{marvosym}

\usepackage{xurl}
\usepackage[colorlinks=true,breaklinks=true]{hyperref}
\usepackage[preprint]{icml2026}

\usepackage{amsmath}
\usepackage{amssymb}
\usepackage{mathtools}
\usepackage{amsthm}

\usepackage[capitalize,noabbrev]{cleveref}

\theoremstyle{plain}

\theoremstyle{definition}

\theoremstyle{remark}

\newcommand{\Benchmark}{MSKernelBench}
\newcommand{\Agent}{CUDAMaster}

\usepackage[textsize=tiny]{todonotes}
\usepackage{threeparttable}
\usepackage[table]{xcolor}
\usepackage{multirow}
\usepackage[breakable]{tcolorbox}
\usepackage{listings}
\definecolor{prompt}{RGB}{59, 130, 246}      
\definecolor{iter0}{RGB}{52, 211, 153}       
\definecolor{iter1}{RGB}{251, 191, 36}       
\definecolor{iter2}{RGB}{168, 85, 247}       
\definecolor{iter3}{RGB}{239, 68, 68}        
\usepackage{graphicx}
\usepackage{subcaption} 
\usepackage[T1]{fontenc}
\usepackage{newtxtext}
\usepackage{newtxmath}
\usepackage{longtable}
\usepackage{makecell}
\usepackage{tabularray}
\usepackage{bm}

\icmltitlerunning{Submission and Formatting Instructions for ICML 2026}

\begin{document}

\twocolumn[
  \icmltitle{
\texorpdfstring{
Making LLMs Optimize Multi-Scenario  CUDA Kernels Like  Experts 
}{
Making LLMs Optimize Multi-Scenario  CUDA Kernels Like  Experts 
}
}

  \icmltitlerunning{Making LLMs Optimize Multi-Scenario  CUDA Kernels Like  Experts 
} 

  \icmlsetsymbol{equal}{*}

  \begin{icmlauthorlist}
    \icmlauthor{Yuxuan Han}{thu,seer}
    \icmlauthor{Meng-Hao Guo}{thu}
    \icmlauthor{Zhengning Liu}{seer}
    \icmlauthor{Wenguang Chen}{thu}
    \icmlauthor{Shi-Min Hu}{thu}

  \end{icmlauthorlist}

  \icmlaffiliation{thu}{Tsinghua University}
  \icmlaffiliation{seer}{Proxseer Inc}


  \icmlkeywords{Machine Learning, ICML}

  \vskip 0.3in
]

\printAffiliationsAndNotice{\ }

\begin{abstract}
Optimizing GPU kernels manually is a challenging and time-consuming task. With the rapid development of LLMs, automated GPU kernel optimization is gradually becoming a tangible reality. However, current LLM-driven automated optimization methods narrowly focus on machine learning applications, such as PyTorch operator optimization, while overlooking broader domains like sparse matrix operations in scientific computing. Extending to these broader applications brings new challenges for the benchmark and algorithm. Therefore, developing a general-purpose automated kernel optimization method becomes our primary focus. In this paper, we address the absence of systematic evaluation for multi-scenario settings by introducing \Benchmark{}, which spans multiple scenarios, including fundamental algebraic operations, common LLM kernels, sparse matrix operators, and scientific computing routines, each supporting both FP32 and BF16 precision. Building on this benchmark, we introduce \Agent{}, a multi-agent, hardware-aware system for kernel optimization that leverages profiling information and automatically constructs the full compilation and execution toolchain. Experimental results demonstrate that \Agent{} achieves significant speedups across most operators, outperforming Astra by about 35\%. In several cases, its performance matches or surpasses that of highly optimized, closed-source libraries such as cuBLAS. A demo showcasing the original and optimized code for each operator is available at \url{https://hanyx2021.github.io/MSKernelBenchDemo/}.

\end{abstract}

\section{Introduction}

\begin{figure*}[t]
    \centering
    \includegraphics[width=0.95\linewidth]{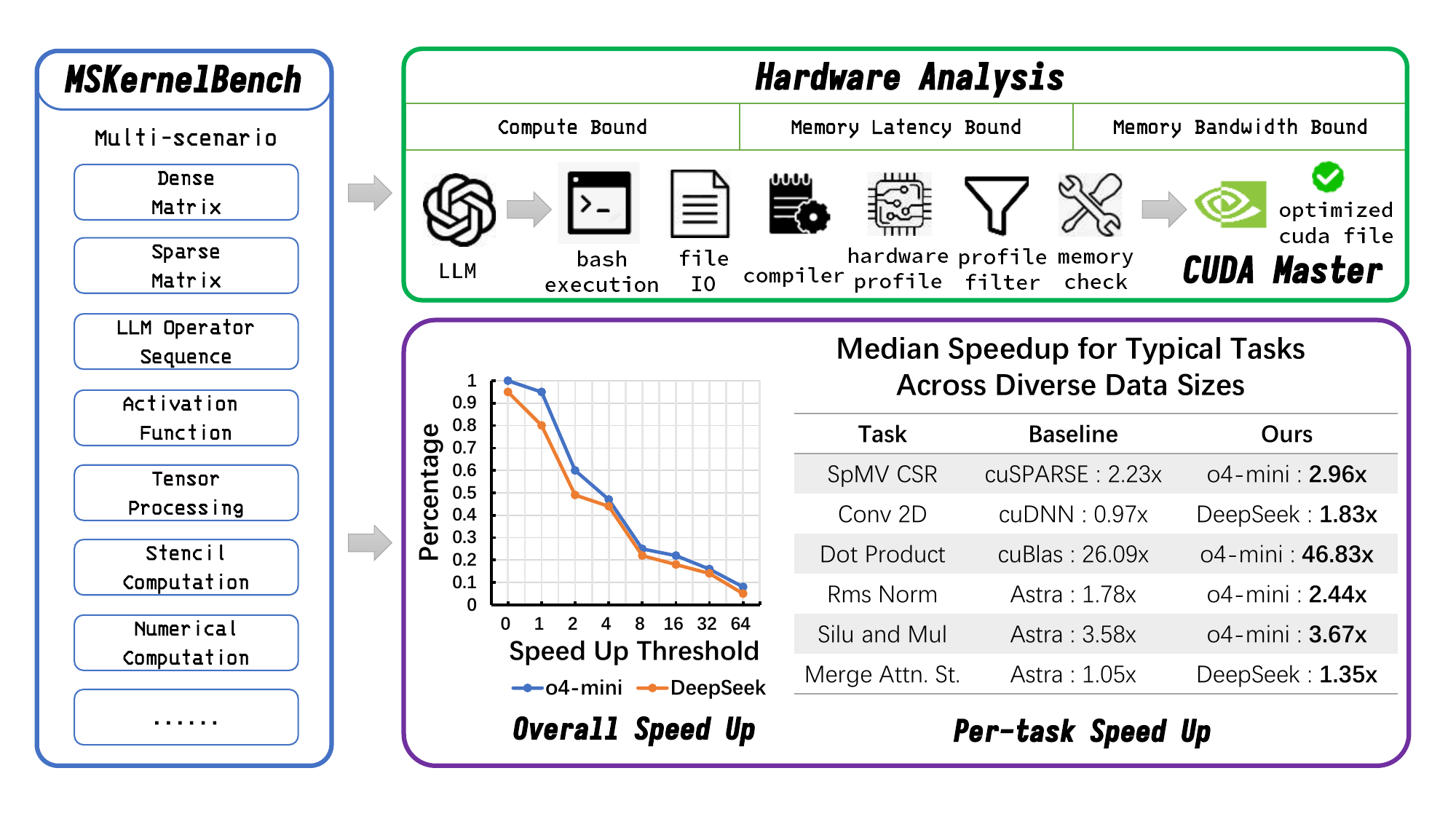}
    \caption{\textbf{Overview of \Benchmark{} and \Agent{}}.\underline{Left}: \Benchmark{} is a comprehensive multi-scenario CUDA optimization benchmark covering dense, sparse, LLM, and scientific kernels, etc. \underline{Top right}: \Agent{} classifies tasks according to performance bottlenecks, filters profile information in a targeted manner, and uses a multi-agent system based on various tools to achieve CUDA kernel optimization. \underline{Bottom right}: On a macro level, the framework successfully balances performance improvements with operator correctness, increasing performance demands across all tested scenarios. On a micro level, the framework can generate kernels surpassing hand-tuned implementations like closed-source libraries across multiple domains.} 
    \label{fig:teaser}
\end{figure*}

Recent advances in Large Language Models (LLMs) have demonstrated remarkable potential in automating various software engineering tasks~\cite{Codegen}. This has naturally spurred interest in applying LLMs to the domain of GPU kernel optimization~\cite{CUDA-LLM}.
Initial explorations, however, have largely been conducted within the context of high-level deep learning frameworks like PyTorch~\cite{Pytorch} and have focused predominantly on a narrow set of operators common in deep learning models, for example KernelBench~\cite{kernelbench}.
While valuable, this approach inherits framework abstractions and optimizations that bias evaluation toward deep-learning-centric execution patterns and fail to capture the broader diversity of GPU kernels found in real-world high-performance computing workloads, such as scientific computing.
Therefore, this work aims to explore general GPU kernel optimization strategies, an area where existing benchmarks and algorithms remain insufficiently developed.

Recent benchmarks, such as KernelBench, have a fundamental limitation in perspective: they equate the problem of \textit{kernel optimization} with the problem of \textit{accelerating LLM components}. These computational tasks are often characterized by computational intensity and regular memory access. This approach fails to test the ability of automated systems to handle more challenging general-purpose computational tasks with more irregular memory access patterns, such as those in sparse linear algebra and scientific computing kernels. Besides, most optimization paths for LLM operators are highly public, and their success may stem from the model's \textit{recall} of known solutions. However, evaluations that include non-LLM operators force the system to \textit{create} in an open question without standard answers, which better tests its true optimization and generalization capabilities.

As for kernel optimization, the multi-scenario optimization task is inherently complex, presenting diverse challenges across sparse linear algebra, scientific simulations, and other domains, each with distinct optimization patterns.
This diversity of patterns leads to difficulties in existing solutions, which are bifurcated: hand-tuned libraries (e.g., cuBLAS~\cite{cublas}, cuSPARSE~\cite{cusparse}) deliver peak performance but lack flexibility and incur immense engineering costs per operator and hardware generation; compiler-based approaches (e.g., TVM~\cite{TVM}, Triton~\cite{Triton}) improve productivity but often struggle to match expert performance across such diverse domains. Therefore, a system capable of automatically optimizing multi-scenario kernels to achieve performance that meets or even surpasses that of closed-source libraries would be a revolutionary advancement. Such a system would guide future library development and solve optimization challenges that even human experts struggle with, thus redefining the upper limit of automated programming powered by LLMs.

In this work, we address it by investigating the limits of LLM-powered agents for general-purpose CUDA kernel optimization.
First, we introduce a comprehensive, multi-scenario benchmark designed to strip away framework dependencies and evaluate optimization capabilities at the fundamental level. It encompasses foundational algebraic operations, common LLM operators, sparse matrix kernels, and scientific computing operators, each supporting FP32 and BF16 data types.
Second, building on it, we propose a multi-agent framework, which strategically leverages selectively filtered hardware profiling data to guide the optimization process and generates complete toolchain requirements.

Our experiments demonstrate that our agent-driven approach achieves significant acceleration on most kernels. Notably, in several cases, it produces code that rivals the performance of heavily optimized, closed-source libraries like cuBLAS and cuSPARSE.
This demonstrates that LLM-based agents, when provided with a suitable environment and information, can approximate expert-level tuning for diverse, low-level GPU programming tasks, paving the way for more adaptive and comprehensive high-performance code generation systems.

In summary, our key contributions are as follows:

\begin{itemize} 
\setlength{\itemsep}{0pt} 

\item 
\textbf{\Benchmark{}}, a comprehensive multi-scenario CUDA optimization benchmark covering dense, sparse, LLM, and scientific kernels, with FP32/BF16 support and multi-scale evaluation.

\item 
\textbf{\Agent{}}, a multi-agent, filtered-profiling-guided, end-to-end optimization framework that generates optimized kernels and the associated toolchain for compilation and execution.

\item  
\textbf{Superior performance.} Experiments show that our method delivers substantial speedups for most operators, outperforming Astra by about 35\%. In some cases, it matches or even exceeds the performance of highly optimized closed-source libraries such as cuBLAS.

\end{itemize}

\section{Related Work}

\subsection{Methodologies for GPU Kernel Optimization}

GPU kernel optimization often follows two paradigms: automated high-level abstractions and manual low-level programming. 
The former leverages compilers and domain-specific languages such as 
TVM~\cite{TVM}
and Triton~\cite{Triton}, which employ auto-tuning (e.g. Ansor~\cite{Ansor}) to explore parameter spaces for development efficiency. 
The latter relies on either hand-tuned vendor libraries such as cuBLAS~\cite{cublas} 
, which set the performance ceiling, or open-source template libraries such as CUTLASS~\cite{cutlass} that provide flexible abstractions for experts. 
Both demand deep hardware expertise and turn each operator into a hardware-dependent engineering challenge.

\subsection{Standardized Benchmarking for Kernel Optimization}

KernelBench~\cite{kernelbench} is established as the authoritative benchmark in this field, systematically evaluating the capability of LLMs to generate performant GPU kernels through 250 PyTorch-based workloads and the \textit{fast\_p} metric. Other notable extensions include Robust‑kbench~\cite{robust-kbench}, which extends by assessing kernel robustness across varied scenarios and data shapes to ensure generalizable improvements; and MultiKernelBench~\cite{MultiKernelBench}, which builds upon KernelBench functionally primarily by adding support for verification across multiple backend platforms. Apart from CUDA, TritonBench ~\cite{TritonBench} is another notable benchmark designed specifically for the Triton GPU programming language.

\subsection{LLM‑Based Optimization of CUDA Kernels}

Recent systems leverage multi‑agent or evolutionary approaches to directly optimize kernels. FSR~\cite{CUDA-LLM} jointly optimizes compilation and runtime performance. CudaForge~\cite{CudaForge} iteratively refines kernels using the profiler feedback from the NVIDIA Nsight Compute tool. Astra~\cite{Astra} uses specialized agents to produce correct, high‑performance kernels selected from SGLang~\cite{SGLang}. In parallel, training‑based methods create specialized models: Kevin~\cite{Kevin} is the first model trained with multi-turn RL for CUDA kernel generation and optimization. CUDA‑L1~\cite{cudaL1} uses contrastive reinforcement learning to learn optimization strategies, while CUDA‑L2~\cite{cudaL2} generates matrix multiplication kernels outperforming cuBLAS~\cite{cublas}.

\begin{table}[t]
  \centering
  \renewcommand{\tabcolsep}{1mm}
  \caption{Comparison of Benchmark Design: KernelBench vs. MSKernelBench.}
  \label{tab:benchmark_compare}
  \makebox[\linewidth][l]{
    \hspace*{-0.4cm}
    \begin{tabular}{lcc}
      \toprule
      \textbf{Feature} & \textbf{KernelBench} & \textbf{\Benchmark{}} \\
      \midrule
      \multicolumn{3}{l}{\textit{Task Scenario Selection}} \\
       Dense matrix operators & $\checkmark$ & $\checkmark$ \\
       LLM operator sequences & $\checkmark$ & $\checkmark$ \\
       Sparse matrix operators & $\times$ & $\checkmark$ \\
       Scientific kernels & $\times$ & $\checkmark$ \\
       Numerical methods & $\times$ & $\checkmark$ \\
      \midrule
      \multicolumn{3}{l}{\textit{Implementation}} \\
       Language & Python & C/C++ \\
       Multi-precision & $\checkmark$ & $\checkmark$ \\
      \midrule
      \multicolumn{3}{l}{\textit{Evaluation Protocol}} \\
       Scalable data sizes & $\times$ & $\checkmark$ \\
       Complexity weighted & $\times$ & $\checkmark$ \\
       Hardware profiling & $\text{limited}$ & $\checkmark$ \\
      \bottomrule
    \end{tabular}
  }
\end{table}

\section{\Benchmark{}: A multi-scenario CUDA operator optimization benchmark}
Current benchmarks for automated CUDA operator tuning focus mainly on common LLM operators and their fused variants, overlooking a wider range of high-performance computing (HPC) scenarios. These tasks have distinct memory, parallelism, and precision requirements. This narrow scope limits the generalizability of tuning research. Additionally, evaluating operators using only a single, fixed data size fails to capture how optimizations scale under varying computational loads and hardware utilization regimes. This limitation undermines the reliability of the reported speedups. Consequently, such a limited scope fails to fully represent the challenges of real-world, high-performance kernel optimization~\cite{METR}.

To address it, we introduce a new benchmark \Benchmark{} designed to support broader, more rigorous evaluation of operator tuning methods, with a summary comparison to KernelBench~\cite{kernelbench} provided in Table \ref{tab:benchmark_compare}.

\subsection{Task selection}

To ensure diversity and reliability, we curated a set of 50 tasks from multiple fields, all having a sizable computational scale. These tasks are collected from authoritative sources, including the official NVIDIA document~\cite{cudasample}, common use cases in CUDA libraries (e.g., cuBLAS, cuSPARSE), open-source benchmarks like KernelBench, and cloud-based CUDA programming learning platform like LeetGPU~\cite{leetGPU}. Detailed task list is provided in the Appendix \ref{app:task_details}.

\subsection{Benchmark Implementation \& Structure}

\subsubsection{Design Rationale: Pure C Language for Portability and Control}

We implemented the benchmark in pure C to prioritize portability and low-level control, both critical for scientific and HPC contexts. C provides a stable and widely supported interface for these domains, facilitating seamless integration with established libraries (e.g., BLAS, sparse solvers) and avoiding the framework-level overhead associated with AI-centric tools like PyTorch~\cite{Pytorch_waste}. Furthermore, C allows fine-grained control over memory access and parallelism, enabling the optimization of irregular computations such as sparse matrix operations.

\subsubsection{Implementation Strategy: Manually Crafted Baselines}

Constructing a comprehensive benchmark in pure C is challenging because well-optimized industrial libraries are usually closed-source. 
Consequently, our strategy is twofold for each task. First, we leverage existing educational or reference implementations where available (e.g., from open-source HPC course materials). Second, when such implementations are unavailable, we manually develop these tasks in C/CUDA. Crucially, every implementation is rigorously validated for numerical correctness through data alignment with a trusted reference, such as a closed-source optimized library (e.g., cuBLAS) or a serial CPU version. Only after passing this validation is it adopted as a performance baseline.

\subsubsection{Benchmark Structure: Precision Separation and Interface Unification}


The benchmark comprises 50 tasks, each implemented in both FP32 and BF16 precision following two design principles: precision separation and interface unification. For precision separation, each operator has distinct FP32 and BF16 implementations, with the latter marked by a "\_bf16" filename suffix. For interface unification, all implementations are accessed through a consistent C wrapper, accompanied by a dedicated test suite built from a common template. This design decouples the core optimization work from the testing harness: researchers can integrate and evaluate custom kernels by replacing only a clearly demarcated optimization segment within the testing template.

\begin{figure}[t]
    \centering
    \includegraphics[width=\linewidth]{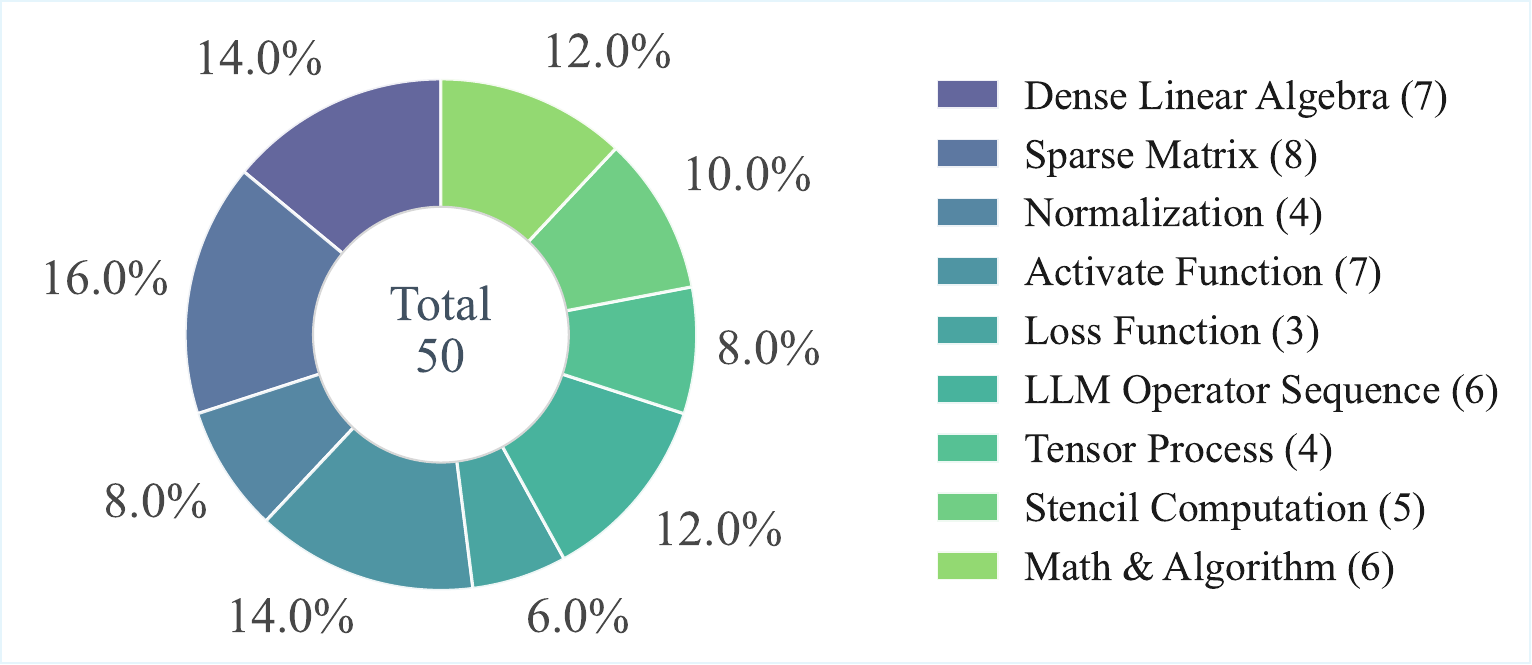}
    \caption{Composition of the 50 Operators by Task Category.} 
    \label{fig:pie_task}
\end{figure}





\subsection{Evaluation metrics}

Building upon the provision of ground-truth kernels, our benchmark implements a rigorous, multi-faceted evaluation protocol consisting of the following metrics:

\textbf{Compilation Correctness} The submitted CUDA code must first be successfully compiled by the test framework. The test harness for each operator enforces a strictly defined C function interface including function name and parameter list. Any deviation will result in a compilation failure, ensuring all subsequent comparisons are based on a consistent calling convention.

\textbf{Multi-Scale Execution Correctness} Each operator is evaluated across a predefined set of data sizes. For each data size, random test cases are generated. Then the output of a submitted optimized kernel is compared against the output of our ground-truth kernel. The submission is correct only if the numerical error is within tolerance across all test scales.

\textbf{Performance Measurement}  Performance is measured as the speedup relative to our provided ground-truth baseline kernel. For each data size, we employ a robust timing methodology: after 3 warm-up runs, the kernel is executed 50 times, and the average execution time is calculated.

The final performance score is computed as a complexity-weighted average of its speedups across all test cases. Formally, the weight $w_i$ for the data size $N_i$ is proportional to the theoretical computational complexity of the baseline implementation (unoptimized) for that input size:
\begin{equation}
w_i \propto T(N_i)
\end{equation}
where $T(N_i)$ represents the theoretical computational complexity of the baseline implementation at data size $N_i$ which is pre-calculated in the test program shown in Appendix \ref{app:task_details}. The overall performance score $P$, designed to represent the average speedup weighed by workload, is defined as:
\begin{equation}
P = \frac{\sum_i w_i S_i}{\sum_i w_i} = \frac{\sum_i T(N_i) S_i}{\sum_i T(N_i)}
\end{equation}
where $S_i$ denotes the measured end-to-end speedup at data size $N_i$ using a consistent and fixed framework.

Since the weights are derived from the complexity of the baseline, larger data sizes naturally receive higher weights. Consequently, if an optimized implementation reduces the actual runtime complexity (e.g., from $O(N^2)$ to $O(N\log N)$), the speedup $S_i$ will be more pronounced for larger $N_i$ . As the larger cases on the scale carry a greater weight on average, the overall score $P$ amplifies the advantage gained from algorithmic improvements, making it a sensitive and equitable metric for comparing optimizations that scale differently with the size of the problem.

\section{\Agent{}: A Multi-agent System Works like Experts}


\subsection{How Human Expert Optimize CUDA Kernels}


Effectively optimizing GPU kernels requires an iterative cycle of implementation, validation, and optimization. This manual process is bottlenecked by two major inefficiencies: laborious data triage and  context-switching workflows. In a conventional workflow, an engineer must manually sift through voluminous hardware profiling data (e.g., from NVIDIA Nsight Compute) to diagnose the specific bottleneck. Extracting the few key performance signatures from overwhelming profiling data is time-consuming and expertise-dependent. Meanwhile, the constant switching between correctness debugging, performance analysis, and code implementation disrupts focus and slows progress. Based on this, our optimization scheme corresponds to two parts: profile filtering and multi-agent cooperation.

\subsection{Hardware Analysis Filter}


To provide targeted profiling guidance, we first establish a taxonomy of hardware bottlenecks. This allows our filter to extract only the relevant metrics for each type. We begin by collecting comprehensive profiling data for all key kernels in our benchmark suite using NVIDIA Nsight Compute. To objectively define the boundaries between bottleneck categories, we apply Otsu's method to the distribution of key throughput metrics across all tasks, determining robust classification thresholds (see Appendix \ref{app:profile_summary}). Combining these thresholds with classical bottleneck characteristics, we classify every task into one of three distinct types:

\textbf{Compute Bound}: High utilization of computational units.

\textbf{Memory Latency Bound}: Significant idle time of computational units due to waiting for data to be ready.

\textbf{Memory Bandwidth Bound}: Saturated utilization of the memory interface bandwidth while computational unit idleness is less pronounced.

Based on this taxonomy, our \textit{hardware analysis filter} operates during execution. For each launched kernel, it leverages a unified mapping table (Table~\ref{tab:bound_metric_mapping}) which defines both the classification criteria for the three bound types and their associated sets of key metrics. The filter classifies the kernel, selects only the relevant metrics according to the table, and discards redundant data. This process distills the profiling context, focusing downstream LLM agents on bottleneck-specific optimization strategies.

\begin{table*}[t]
\caption{Bottleneck Classification and Metric Filtering Strategy}
\label{tab:bound_metric_mapping}
\centering
\begin{tabular}{c  >{\centering\arraybackslash}m{3.5cm} >{\centering\arraybackslash}m{7.5cm}}
\toprule
\textbf{Bottleneck Type} & \textbf{Classification Rule} & \textbf{Filtered Metrics} \\
\midrule
Compute Bound & \text{Compute.Th} > 30\% & Compute (SM) Throughput, Issue Slots Busy, Executed Ipc Active, SM Busy \\
\addlinespace[0.3em]
\midrule
Memory Latency Bound & \text{Compute.Th} < 30\%, \newline \text{DRAM.Th} < 30\% and \newline \text{Memory.Th} < 30 \% & L2 Hit Rate, L1/TEX Hit Rate, \newline Executed Ipc Elapsed, Mem Busy  \\
\midrule
\addlinespace[0.3em]
Memory Bandwidth Bound & \text{Compute.Th} < 30\%, \newline \text{DRAM.Th} > 30\% or \newline \text{Memory.Th} > 30 \% & DRAM Throughput, Memory Throughput, \newline Max Bandwidth, Mem Pipes Busy \\
\bottomrule
\end{tabular}
\end{table*}

\begin{figure*}[t]
    \centering
    \includegraphics[width=0.9\linewidth]{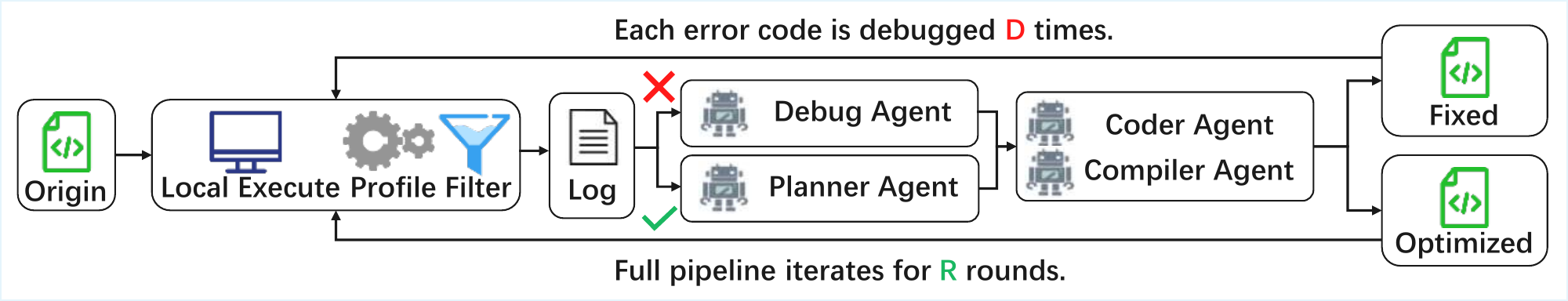}
    \caption{\Agent{} Framework for Automated Kernel Optimization Pipeline.}
    \label{fig:pipeline}
\end{figure*}

\begin{algorithm}[t]
\caption{\Agent{} Pipeline}
\label{alg:framework}
\begin{algorithmic}[1]
\INPUT Number of iterations $R$, max debug rounds $D$, \\
original kernel code $C_0$, original execution script $S_0$
\OUTPUT Best code $C^*$, best execution script $S^*$
\STATE Record profile $\pi_0 \gets \text{ExecuteAndFilter} (C_0,S_0)$
\STATE Initialize $C^*\gets C_0$, $S^*\gets S_0$, $P^* \gets 1.0$, \\
$\pi^*\gets\pi_0$, scheme $\gets$ "Baseline"
\FOR{$i = 1$ to $R$}
    \STATE $\text{scheme} \gets \text{PlannerAgent}(C^*,\pi^*,\text{scheme})$

    \STATE $C_i \gets \text{CoderAgent}(\text{scheme}, C^*)$
    \STATE $S_i \gets \text{CompilerAgent}(\text{scheme}, C_i)$
    \STATE $\pi_i \gets \text{ExecuteAndFilter}(C_i, S_i)$
    
    \IF{$\text{CheckValid}(\pi_i) = \text{False}$}
        \FOR{$d = 1$ to $D$}
            \STATE $\text{debug\_scheme} \gets \text{DebugAgent}(C^*,\pi^*)$
            \STATE $C_i^{(d)} \gets \text{CoderAgent}(\text{debug\_scheme}, C_i)$
            \STATE $S_i^{(d)} \gets \text{CompilerAgent}(\text{scheme}, C_i^{(d)})$
            \STATE $\pi_i^{(d)} \gets \text{ExecuteAndFilter}(C_i^{(d)}, S_i^{(d)})$
                \IF{$\text{CheckValid}(\pi_i^{(d)}) = \text{True}$} 
                    \STATE $(C_i,S_i,\pi_i) \gets (C_i^{(d)},S_i^{(d)},\pi_i^{(d)})$
                    \STATE \textbf{BREAK}
                \ENDIF
        \ENDFOR
    \ENDIF
    \IF{$\text{CheckValid}(\pi_i) = \text{True} \  \AND P(\pi_i) > P^*$}
        \STATE $(C^*,S^*,\pi^*) \gets (C_i,S_i,\pi_i)$
        \STATE Update scheme
    \ELSE
        \STATE Mark scheme as invalid
    \ENDIF
\ENDFOR
\STATE \textbf{return} $C^*,S^*$
\end{algorithmic}
\end{algorithm}

\subsection{\Agent{}: A Multi-agent System}

As illustrated in Figure \ref{fig:pipeline}, we propose a multi-agent framework consisting of four specialized agents to automate the kernel tuning process. The complete execution flow of this system, governed by Algorithm \ref{alg:framework}, is controlled by two key parameters: the number of iteration rounds $R$ and the maximum number of debugging rounds $D$.

The process begins by running the original kernel to collect a performance baseline and hardware profiling data, which is filtered to isolate key bottleneck metrics. The system then iterates for $R$ rounds. In each round, the \textit{Planner Agent} first analyzes the distilled profiling insights from the historical data to propose high-level optimization strategies. These strategies are then passed to the \textit{Coder Agent}, which implements them into executable kernel code within the provided test harness. Subsequently, the \textit{Compiler Agent} manages all compilation commands, execution scripts, and compiler-level optimizations to build and run the code. If the newly generated kernel fails the execution correctness check, a debugging subprocess (limited to $D$ rounds) is invoked. Within this subprocess, the \textit{Debug Agent} is responsible for diagnosing and correcting any errors encountered during code implementation or compilation, ensuring the robustness of the tuning process. Successful kernels are evaluated, and the best-performing solution is retained, enabling continuous exploration of the optimization space.

Throughout this iterative process, the entire testing and execution environment remains anchored to the benchmark's standardized code. None of the agents alter the core test logic, guaranteeing that every optimized kernel is evaluated under consistent and fair conditions.

\section{Experiment}

\subsection{Experimental Setup}

We evaluate our method using \Benchmark{} on 50 distinct tasks, each implemented in both FP32 and BF16 precision, resulting in a total of 100 kernel optimization tasks.

\subsubsection{Base LLMs \& Hardware}
We evaluate our framework on two leading LLMs in code generation: OpenAI o4-mini and DeepSeek-V3.2~\cite{deepseek}. All kernel compilation and performance profiling are conducted on an NVIDIA RTX 4090 GPU.

\subsubsection{Agent Configuration}
Our framework uses the following fixed parameters: iteration rounds $R = 3$ and debug trials per candidate $D = 3$.

\subsubsection{Metrics}

We use two complementary metrics: a macro-analysis of cumulative success across all tasks, and a micro-analysis of per-operator speedup against established baselines.

\textbf{Cumulative Success Rate (Macro-analysis)}: We define the cumulative success rate as the percentage of tasks, whose complexity-weighted average speedup $P$ exceeds a threshold $\tau$. We focus mainly on results at key thresholds: $\tau=0$ (functional correctness), $\tau=1$ (beating the naive baseline), and $\tau>1$ (degree of optimization).

\textbf{Per-Operator Speedup (Micro-analysis)}: To gauge precise performance, we evaluate a subset of tasks that have authoritative baselines: either highly-optimized closed-source libraries (e.g., cuBLAS, cuSPARSE) or state-of-the-art research implementations (e.g., Astra~\cite{Astra} for optimizing SGLang~\cite{SGLang} kernels). For each task, we measure the speedup of our generated implementation against the naive baseline provided by \Benchmark{}. We then compare this value directly with the speedup achieved by the authoritative baseline over the same naive baseline. This reveals our method's absolute effectiveness and competitiveness compared with prior work.

\begin{figure}[t]
    \centering
    \includegraphics[width=\linewidth]{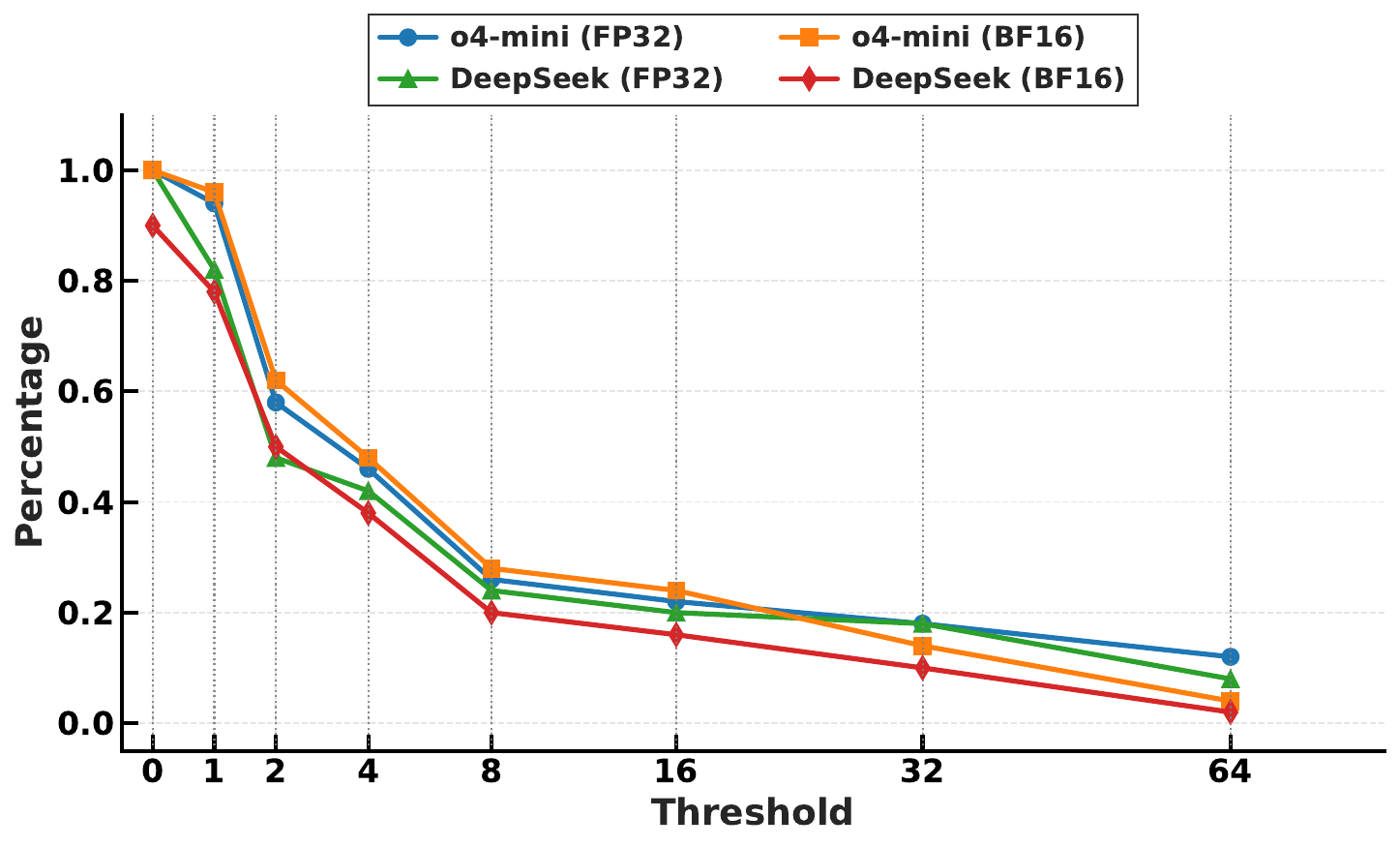}

    \caption{
    Cumulative success rate of all tasks versus speedup threshold($\tau$) for OpenAI o4-mini and DeepSeek-V3.2 models under FP32 and BF16 precision.
    }
    \label{fig:cumulative}
\end{figure}

\subsection{Results on Cumulative Success Distribution}

As shown in Figure \ref{fig:cumulative}, o4-mini demonstrates a clear and consistent performance advantage over DeepSeek-V3.2 in generating successful optimizations across speedup thresholds ($\tau$) and numerical precisions (FP32/BF16). This lead is substantial at key thresholds: 100\% vs. 95\% at $\tau=0$, 94\% vs. 80\% at $\tau=1$, and 60\% vs. 49\% at $\tau=2$.

The effect of numerical precision is secondary but notable: BF16 offers a slight edge for o4-mini at moderate-to-high thresholds ($\tau \leq 8$), while FP32 ensures greater stability for both models under stringent optimization targets ($\tau \geq 32$).

These results confirm that our optimization framework effectively balances correctness with performance gains, with o4-mini proving to be the more robust model for this task.

\subsection{Results on Single Task Speedup Compared with Established Implementation}

\begin{figure*}[t]
    \centering
\begin{subfigure}{0.33\textwidth}
        \centering
        \includegraphics[width=\linewidth]{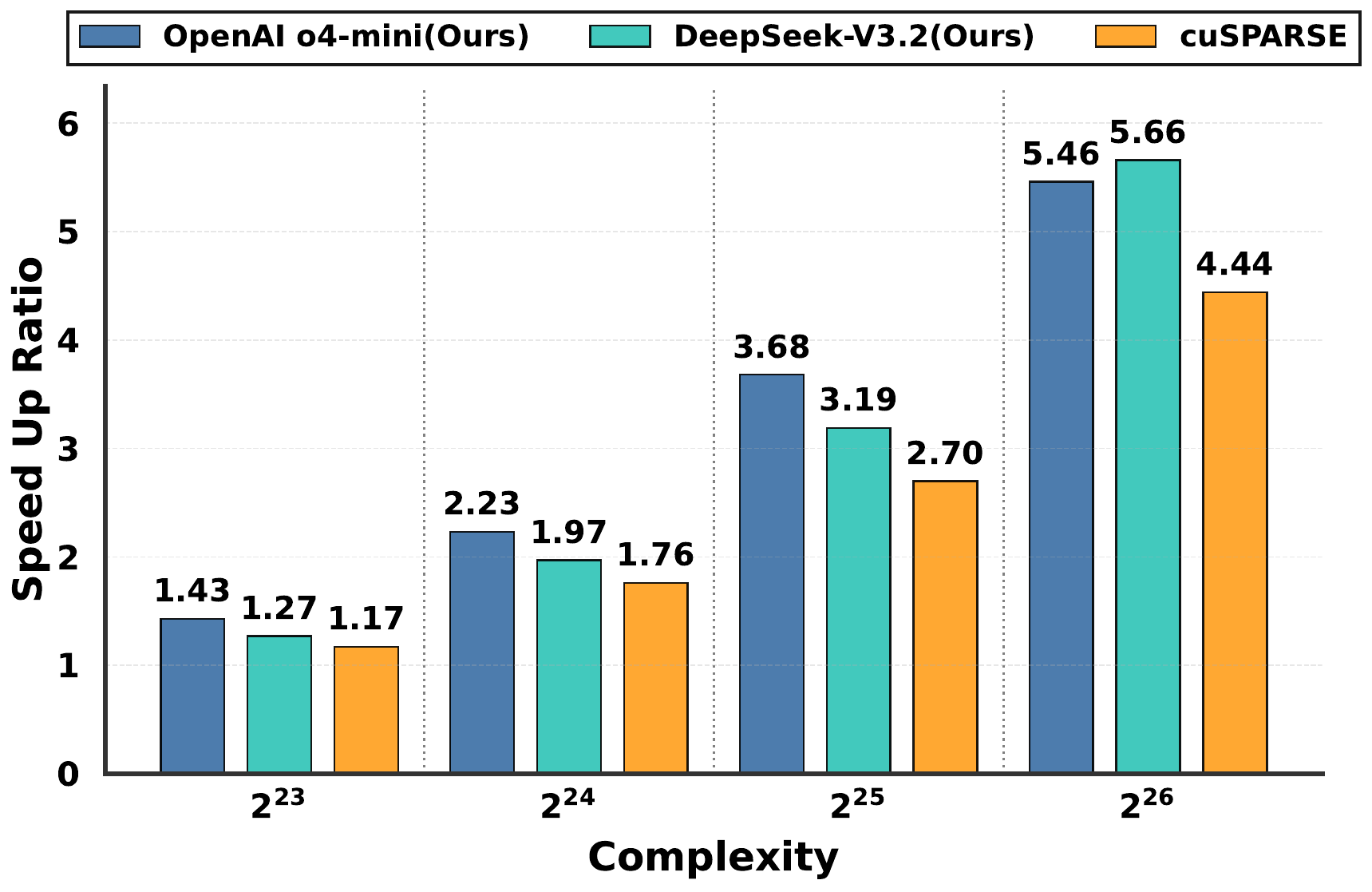}
        \caption{SpMV CSR vs. cuSPARSE}
        \label{fig:spmv_csr_single}
    \end{subfigure}\hfill%
    \begin{subfigure}{0.33\textwidth}
        \centering
        \includegraphics[width=\linewidth]{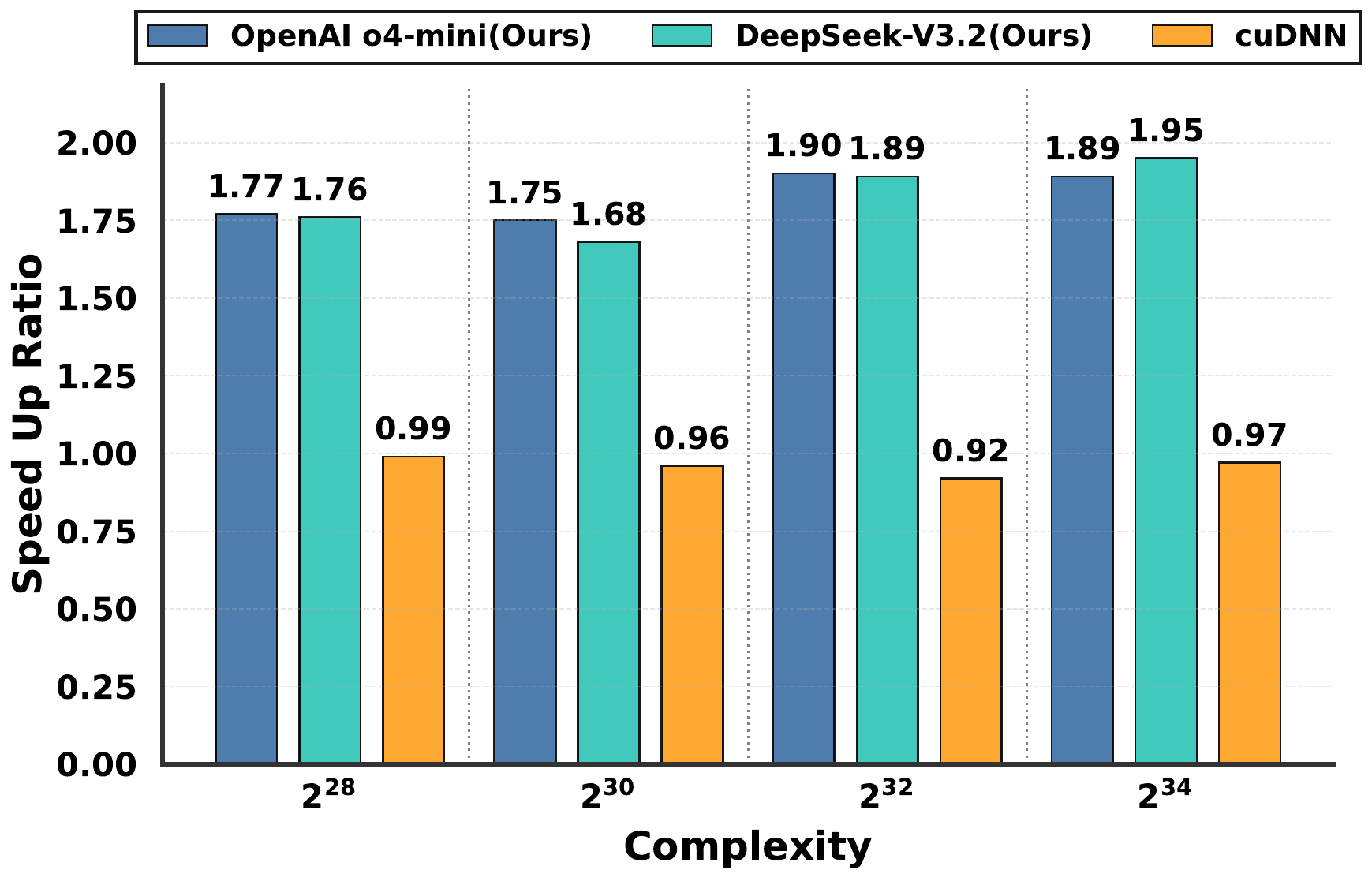}
        \caption{2D Convolution vs. cuDNN}
        \label{fig:conv_2d_single}
    \end{subfigure}\hfill%
    \begin{subfigure}{0.33\textwidth}
        \centering
        \includegraphics[width=\linewidth]{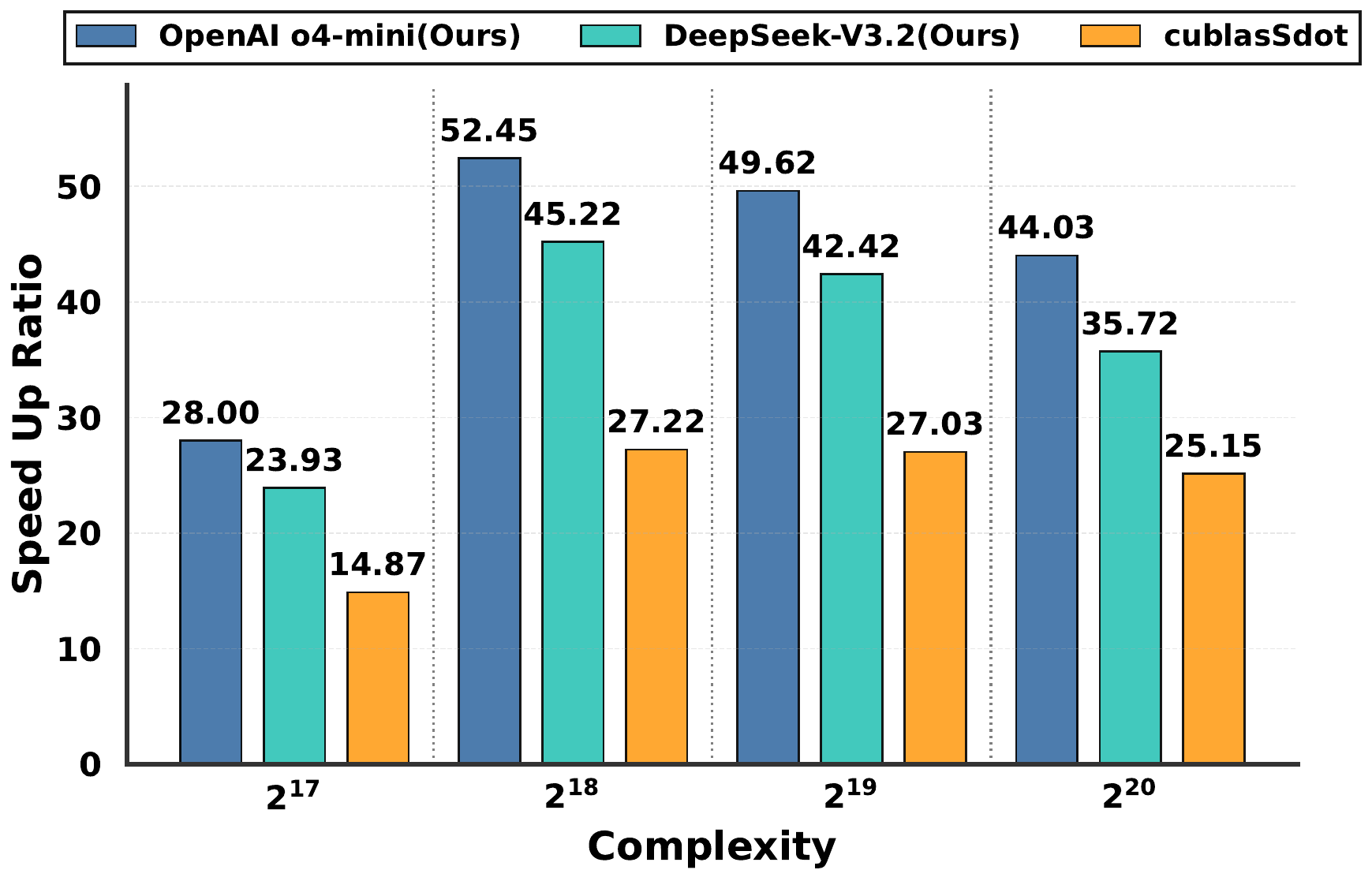}
        \caption{Dot Product vs. cuBLAS}
        \label{fig:dot_product_single}
    \end{subfigure}

    \begin{subfigure}{0.33\textwidth}
        \centering
        \includegraphics[width=\linewidth]{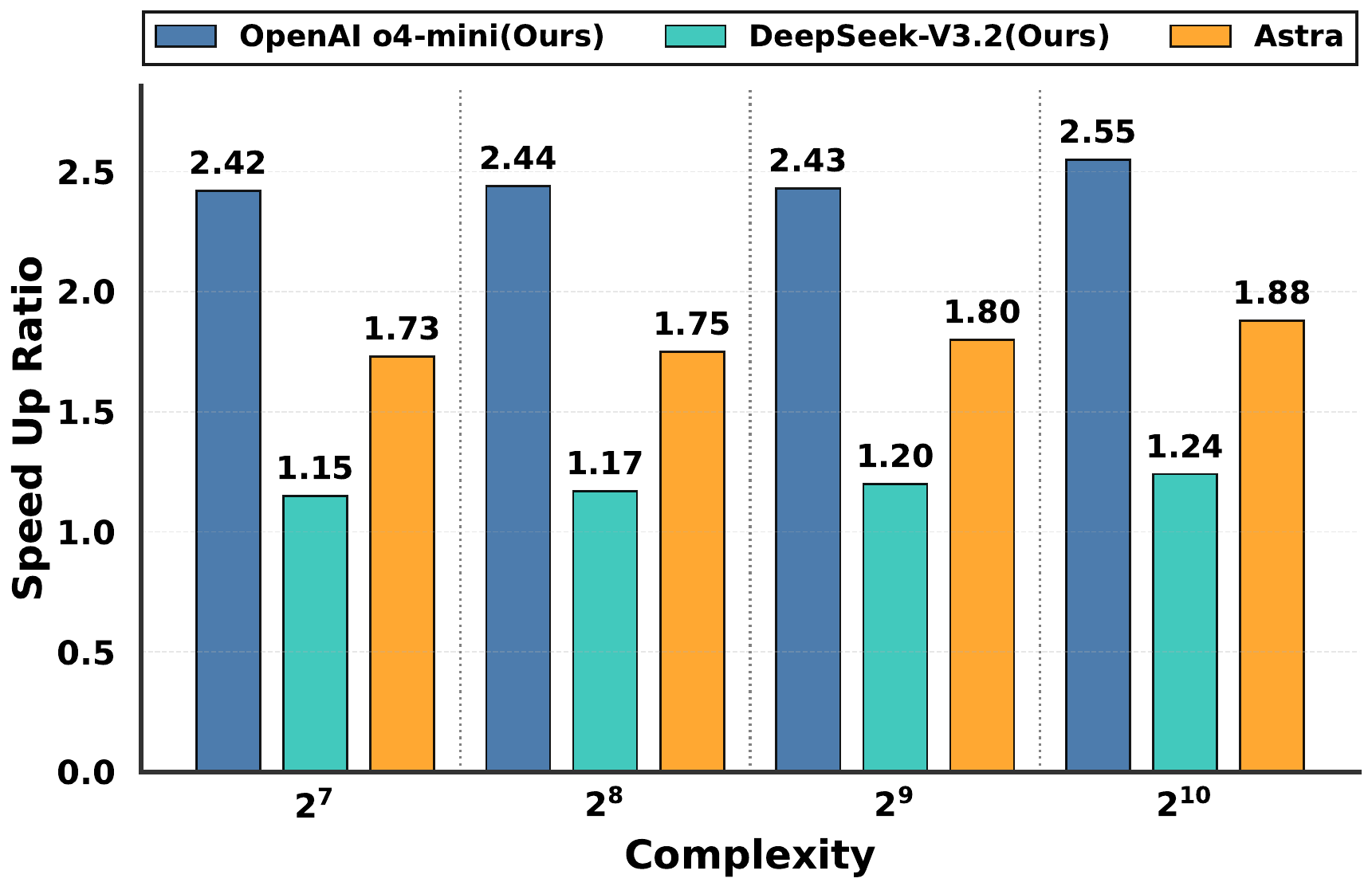}
        \caption{RMSNorm vs. Astra}
        \label{fig:rms_norm}
    \end{subfigure}
    \begin{subfigure}{0.33\textwidth}
        \centering
        \includegraphics[width=\linewidth]{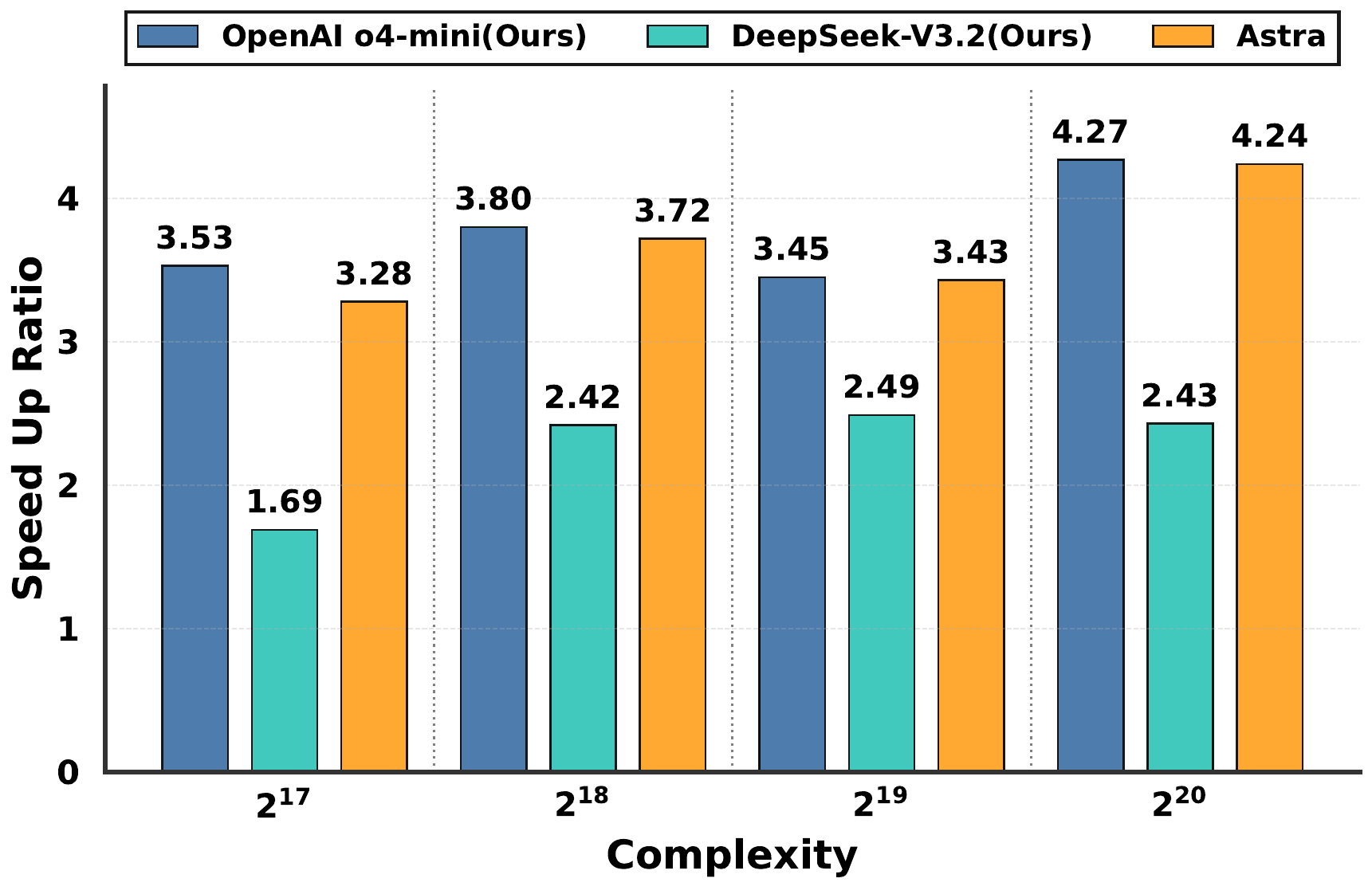}
        \caption{SiLU\&Mul vs. Astra}
        \label{fig:silu_mul}
    \end{subfigure}
    \begin{subfigure}{0.33\textwidth}
        \centering
        \includegraphics[width=\linewidth]{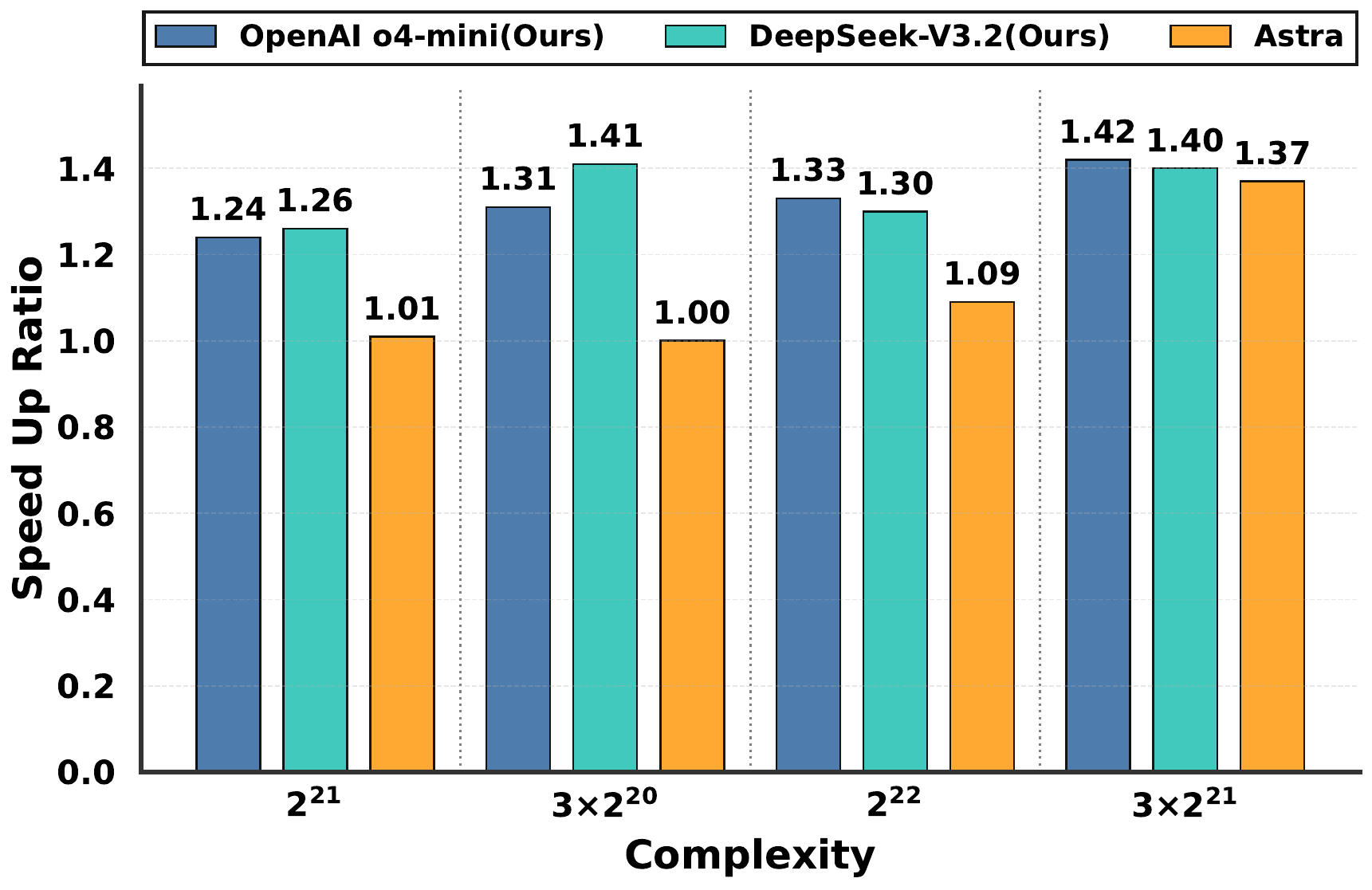}
        \caption{Merge Attention States vs. Astra}
        \label{fig:merge_attn_states}
    \end{subfigure}
    
    \caption{
    Performance comparison of our agent-optimized kernels against closed-source libraries (cuBLAS, cuDNN, cuSPARSE) and existing optimization frameworks (Astra) across six representative operators and varying problem scales.
    Subfigures (a), (b), and (c) show performance advantages over closed-source libraries in three key operations: (a) sparse matrix-vector multiplication (SpMV) over cuSPARSE, (b) 2D convolution over cuDNN, and (c) dot product over cuBLAS.
    Subfigures (d), (e), and (f) show competitive performance against Astra on LLM operators (RMSNorm, SiLU\&Mul and Merge Attention States).
    }
    \label{fig:library_comparison}
\end{figure*}

We evaluated our method on six representative operators with authoritative baselines: three from industry-standard libraries (cuBLAS: Dot Product, cuDNN~\cite{cudnn}: 2D Convolution, cuSPARSE: SpMV CSR) and three from SGLang~\cite{SGLang} kernel optimized by Astra~\cite{Astra} framework (RMS norm, SiLU \& Mul, Merge Attention States). For each, we compare the speedup achieved by our agent with that of the baseline relative to a common reference.

As shown in Figure \ref{fig:library_comparison}, our framework delivers competitive or superior performance across all tested cases. Both models outperform cuSPARSE in SpMV CSR and achieve up to 1.8$\times$ speedup over cuDNN in 2D Convolution and over cuBLAS in Dot Product. Against the Astra-optimized kernels, o4-mini leads by around 35\% on RMSNorm and matches or surpasses it on the other two fused operators.

These results validate that automated optimization can not only match but also exceed expert-hand-tuned libraries, with o4-mini demonstrating particular strength in both general-purpose and specialized computations.

\subsection{Ablation Study}

\begin{table}[t]
\centering
\footnotesize
\caption{Cumulative success rate across all multi-agent parameters.}
\label{tab:R_and_D_ablation}
\begin{tabular}{cccccc}
\toprule
\multirow{2}{*}{\textbf{Model}} & \multirow{2}{*}{\textbf{$\tau$}} & \multicolumn{4}{c}{\textbf{parameters}} \\
\cmidrule(lr){3-6}
 & & \textbf{Full} & \textbf{N.D.} & \textbf{S.I.} & \textbf{S.R.} \\
\midrule

\multirow{5}{*}{OpenAI o4-mini}
 & 0 & \textbf{100\%} & 90\% & 96\% & 90\% \\
 & 1 & \textbf{94\%} & 77\% & 74\% & 77\% \\
 & 2 & \textbf{60\%} & 50\% & 46\% & 50\% \\
 & 4 & \textbf{47\%} & 35\% & 35\% & 35\% \\
 & $\geq$8 & \textbf{25\%} & 19\% & 17\% & 19\% \\
 
\addlinespace[0.2em]

\midrule

\multirow{5}{*}{DeepSeek-V3.2}
 & 0 & \textbf{95\%} & 87\% & 82\% & 82\% \\
 & 1 & \textbf{80\%} & 67\% & 61\% & 61\% \\
 & 2 & \textbf{49\%} & 36\% & 37\% & 37\% \\
 & 4 & \textbf{40\%} & 30\% & 31\% & 31\% \\
 & $\geq$8 & \textbf{22\%} & 15\% & 18\% & 18\% \\

\bottomrule
\end{tabular}
\begin{flushleft}\footnotesize 
\emph{Note}: Abbreviations: N.D. (No Debug), S.I. (Single Iteration), S.R. (Single Run).
\end{flushleft}
\end{table}

\begin{table}[t]
\centering
\footnotesize
\caption{Cumulative success rate across all profile patterns.}
\label{tab:filter_ablation}
\begin{tabular}{cccccc}
\toprule
\multirow{2}{*}{\textbf{Model}} & \multirow{2}{*}{\textbf{$\tau$}} & \multicolumn{3}{c}{\textbf{Profile Pattern}} \\
\cmidrule(lr){3-5}
 & & \textbf{Filtered} & \textbf{No Prof.} & \textbf{Full Prof.} \\
\midrule

\multirow{5}{*}{OpenAI o4-mini}
 & 0 & \textbf{100\%} & 99\% & 100\% \\
 & 1 & \textbf{94\%} & 90\% & 94\% \\
 & 2 & 60\% & 56\% & \textbf{62\%} \\
 & 4 & \textbf{47\%} & 43\% & 47\% \\
 & $\geq$8 & \textbf{25\%} & 22\% & 25\% \\
 
\addlinespace[0.2em]

\midrule

\multirow{5}{*}{DeepSeek-V3.2}
 & 0 & \textbf{95\%} & 95\% & 94\% \\
 & 1 & \textbf{80\%} & 69\% & 73\% \\
 & 2 & \textbf{49\%} & 41\% & 43\% \\
 & 4 & \textbf{40\%} & 34\% & 33\% \\
 & $\geq$8 & \textbf{22\%} & 20\% & 16\% \\

\bottomrule
\end{tabular}
\end{table}

\begin{table}[t]
\centering
\footnotesize
\caption{Average cost and token usage per task across all profile patterns.}
\label{tab:cost_token_ablation}
\begin{tabular}{lcc}
\toprule
\textbf{Configuration} & \textbf{Average Cost} & \textbf{Average Tokens} \\
\midrule
\textbf{OpenAI o4-mini} \\
\quad Filtered Profile & \$0.2707 & 123,036 \\
\quad No Profile & \$0.2120 & 96,373 \\
\quad Full Profile & \$0.3804 & 172,917 \\
\addlinespace[0.3em]
\hline
\addlinespace[0.3em]
\textbf{DeepSeek-V3.2} \\
\quad Filtered Profile & \$0.0310 & 223,294 \\
\quad No Profile & \$0.0277 & 167,971 \\
\quad Full Profile & \$0.0687 & 387,432 \\
\bottomrule
\end{tabular}
\end{table}

\subsubsection{Impact of Iterative Planning and Debugging}

We conducted an ablation study on our framework's two key parameters: iteration rounds ($R$) and debugging attempts per candidate ($D$). We evaluated four configurations: Full ($R=3$, $D=3$), No Debug (N.D.) ($D=0$), Single Iteration (S.I.) ($R=1$) and Single Run (S.R.) ($R=1$, $D=0$).

Results in Table \ref{tab:R_and_D_ablation} show that the Full configuration consistently achieves the highest success rates. For instance, using the o4-mini model at $\tau=1$, it attains a 94\% success rate, surpassing the N.D. (77\%) and S.I. (74\%) configurations. The performance gap widens at more stringent thresholds ($\tau \geq 8$), where Full maintains a 25\% success rate, compared to 17–19\% for the ablated variants. The Single Run baseline performs the worst, highlighting the necessity of a structured optimization process. Experiments with DeepSeek-V3.2 exhibit a similar trend, albeit at an overall lower performance level. These results demonstrate that both iterative refinement and systematic debugging are indispensable components of our framework for achieving substantial speedups in challenging optimization scenarios.

\subsubsection{Impact of Profile-Guided Optimization}

We further analyze the effect of hardware profiling strategies through three configurations: Filtered, No Profile (No Prof.), and Full Profile (Full Prof.). Results in Table \ref{tab:filter_ablation} demonstrate the effectiveness of our filtered approach.

For o4-mini, the Filtered configuration matches Full Profile performance across all thresholds while reducing profiling overhead. At $\tau=1$, both achieve 94-95\% success, higher than No Profile (90\%). This advantage persists at higher thresholds, with Filtered maintaining 25\% success at $\tau\geq8$ versus No Profile's 22\%. DeepSeek-V3.2 shows similar trends, though with more variation between configurations. These results validate our filtered profiling strategy as an efficient mechanism for providing targeted guidance.

Furthermore, analysis of API cost and token consumption in Table \ref{tab:cost_token_ablation} shows that our filtering strategy provides an optimal trade-off. It retains the critical hardware diagnostics that the low-cost No-Profile configuration lacks, while significantly reducing the resource overhead of the Full-Profile approach—cutting cost by up to 32\% and token usage by 30–40\%. This confirms that bottleneck-aware filtering effectively controls the expense of LLM-assisted optimization without sacrificing optimization guidance.

\section{Conclusion}

We introduce \Benchmark{}, a comprehensive and diverse suite of operators spanning multiple domains and numerical precisions, establishing a rigorous testbed for GPU code optimization. Building on this benchmark, we propose \Agent{}, an end-to-end multi-agent framework designed to interpret hardware profiling data and autonomously generate, debug, and deploy optimized CUDA code alongside its complete toolchain. Experimental results demonstrate that our agent-driven approach achieves significant speedups across this benchmark and produces code competitive with manually tuned libraries in several cases.

Overall, this work demonstrates that LLM-based agents can effectively automate the optimization of complex, low-level CUDA kernels, achieving significant speedups across a diverse set of scenarios. By open-sourcing our framework and benchmark, we provide a foundation for making LLMs optimize multi-scenario CUDA kernels like experts.



\bibliography{example_paper}
\bibliographystyle{icml2026}

\appendix
\onecolumn

\section{Hardware Profile Summary}
\label{app:profile_summary}

\subsection{Full Metrics}
We employ NVIDIA Nsight Compute (v2024.1) to collect fine-grained hardware performance counters. Table~\ref{tab:ncu-config} lists the key metrics we filter and extract from these counters, organized by profiling section:

\begin{table}[h]
\centering
\caption{Filtered Hardware Performance Metrics for Bottleneck Analysis by Profiling Section.}
\label{tab:ncu-config}
\small
\setlength{\extrarowheight}{2pt}
\begin{tabular}{c >{\centering\arraybackslash}m{3.5cm} >{\centering\arraybackslash}m{7.5cm}}
\toprule
\textbf{Profiling Section} & \textbf{Analysis Focus} & \textbf{Filtered Key Metrics} \\
\addlinespace[2pt] 
\midrule
\texttt{Compute Workload Analysis} & Computational Efficiency & Executed Ipc Active, Executed Ipc Elapsed, Issue Slots Busy, Issued Ipc Active, SM Busy \\
\addlinespace[2pt] 
\hline
\texttt{GPU Speed Of Light} & Throughput \& Utilization & Compute (SM) Throughput, DRAM Throughput, Duration, Memory Throughput \\
\addlinespace[2pt] 
\hline
\texttt{Memory Workload Analysis} & Memory Subsystem & L1/TEX Cache Throughput, L1/TEX Hit Rate, L2 Cache Throughput, L2 Hit Rate, Max Bandwidth, Mem Busy, Mem Pipes Busy \\
\addlinespace[2pt] 
\hline
\texttt{Occupancy} & Hardware Resource Usage & Achieved Active Warps Per SM, Achieved Occupancy \\
\addlinespace[2pt] 
\hline
\texttt{Scheduler Statistics} & Instruction Scheduler & Active Warps Per Scheduler, Eligible Warps Per Scheduler, Issued Warp Per Scheduler, No Eligible, One or More Eligible \\
\addlinespace[2pt] 
\hline
\texttt{Warp State Statistics} & Warp Execution & Warp Cycles Per Executed Instruction, Warp Cycles Per Issued Instruction \\
\addlinespace[2pt] 
\bottomrule
\end{tabular}
\end{table}

\subsection{Bottleneck Classification Logic}
\label{app:classification_logic}

To translate raw profiling data into actionable optimization guidance, our system first identifies the primary performance bottleneck. This classification is based on the analysis of four foundational metrics:

\textbf{Duration}: Provides the absolute execution time, serving as the baseline for measuring the impact of any optimization.

\textbf{Compute (SM) Throughput}: Indicates the utilization level of the streaming multiprocessors (SMs).

\textbf{DRAM Throughput}: Reflects the utilization of the GPU's main memory interface, the most critical indicator for bandwidth saturation.

\textbf{Memory Throughput}: Represents the total data movement throughput across the entire memory hierarchy (caches and DRAM).

The three throughput metrics \textbf{Compute (SM) Throughput}, \textbf{DRAM Throughput} and \textbf{Memory Throughput} serve as the primary axes for our bottleneck classification. To move beyond heuristic thresholds and establish objective criteria, we adopted a data-driven calibration process.

For each operator in our benchmark, we first identified its most time-consuming, performance-critical bottleneck CUDA kernel via \textbf{Duration}. We then executed all baseline (unoptimized) implementations of these kernels and collected three key throughput metrics. This process resulted in a dataset where each data point corresponds to the throughput profile of a distinct bottleneck kernel.

To determine the optimal threshold that separates “high” and “low” utilization for each throughput metric, we applied \textbf{Otsu's method} – a classical algorithm for image thresholding that maximizes the inter-class variance. In our context, it automatically finds a threshold value that best separates the distribution of a given metric into two classes, effectively identifying the value where the distinction between “compute-bound” and “non-compute-bound” kernels is most pronounced.

\begin{figure*}[h]
    \centering
    \begin{subfigure}{0.33\textwidth}
        \centering
        \includegraphics[width=\linewidth]{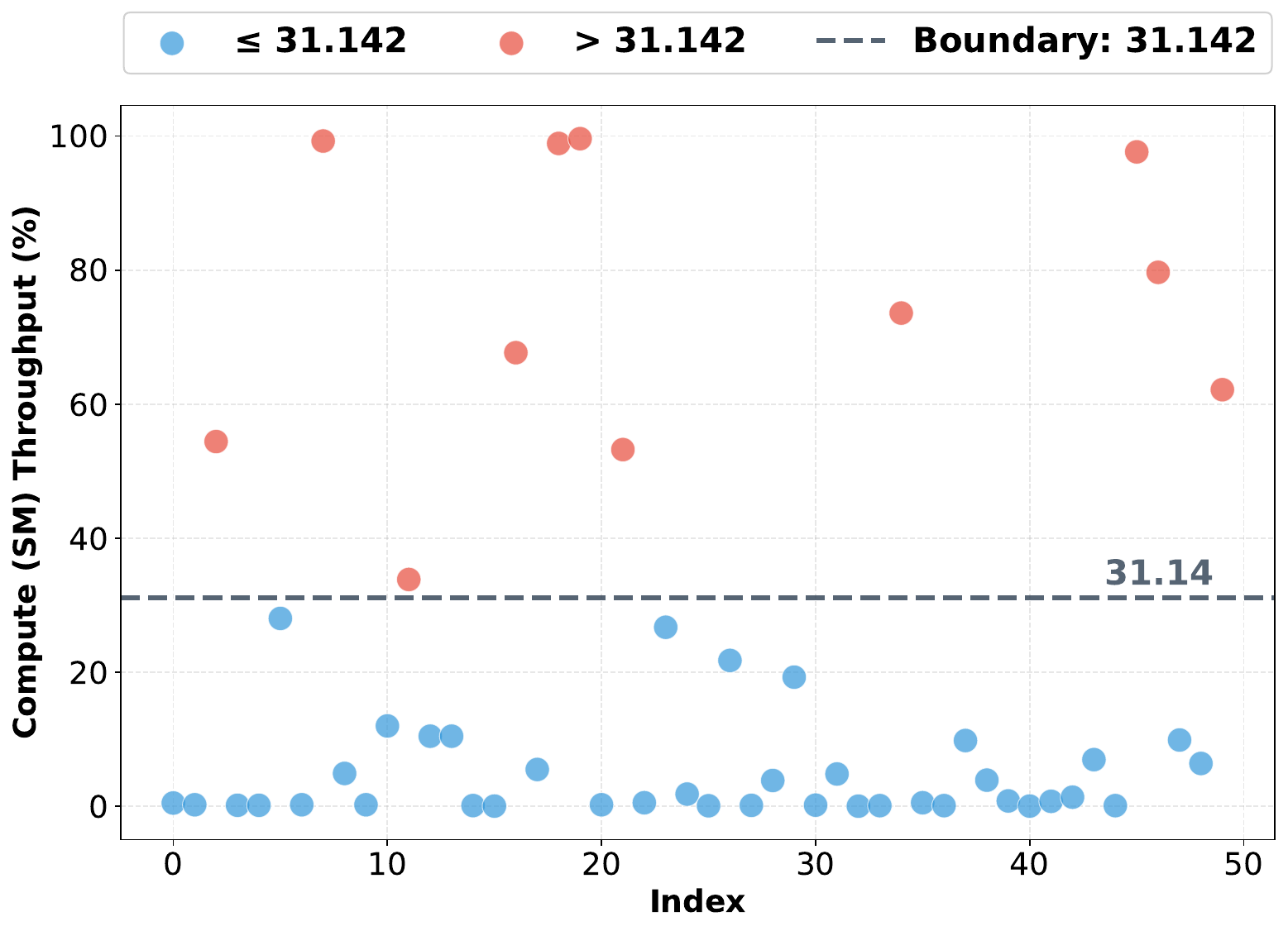}
        \caption{Compute (SM) Throughput}
        \label{fig:scatter_compute}
    \end{subfigure}\hfill%
    \begin{subfigure}{0.33\textwidth}
        \centering
        \includegraphics[width=\linewidth]{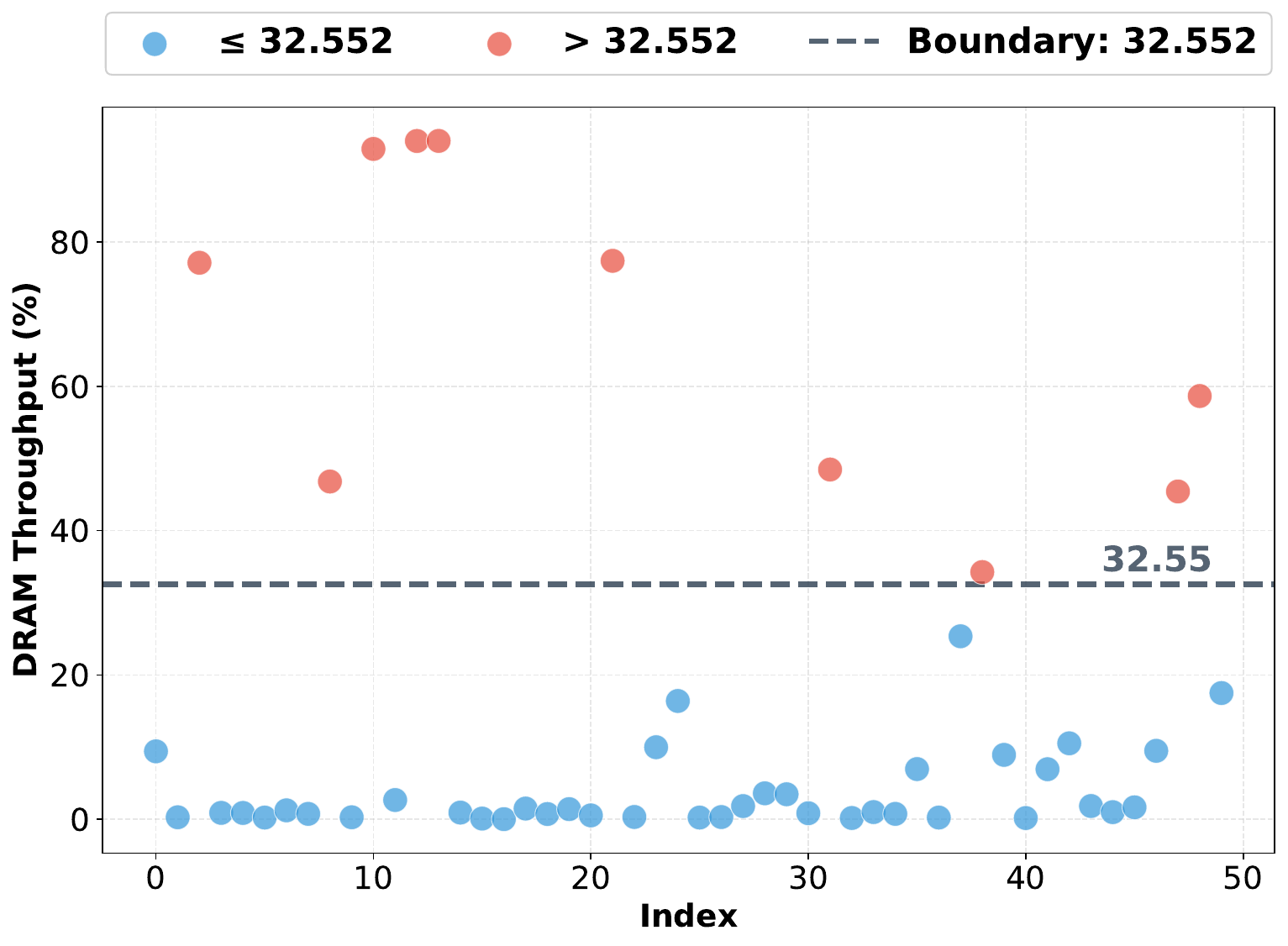}
        \caption{DRAM Throughput}
        \label{fig:scatter_dram}
    \end{subfigure}\hfill%
    \begin{subfigure}{0.33\textwidth}
        \centering
        \includegraphics[width=\linewidth]{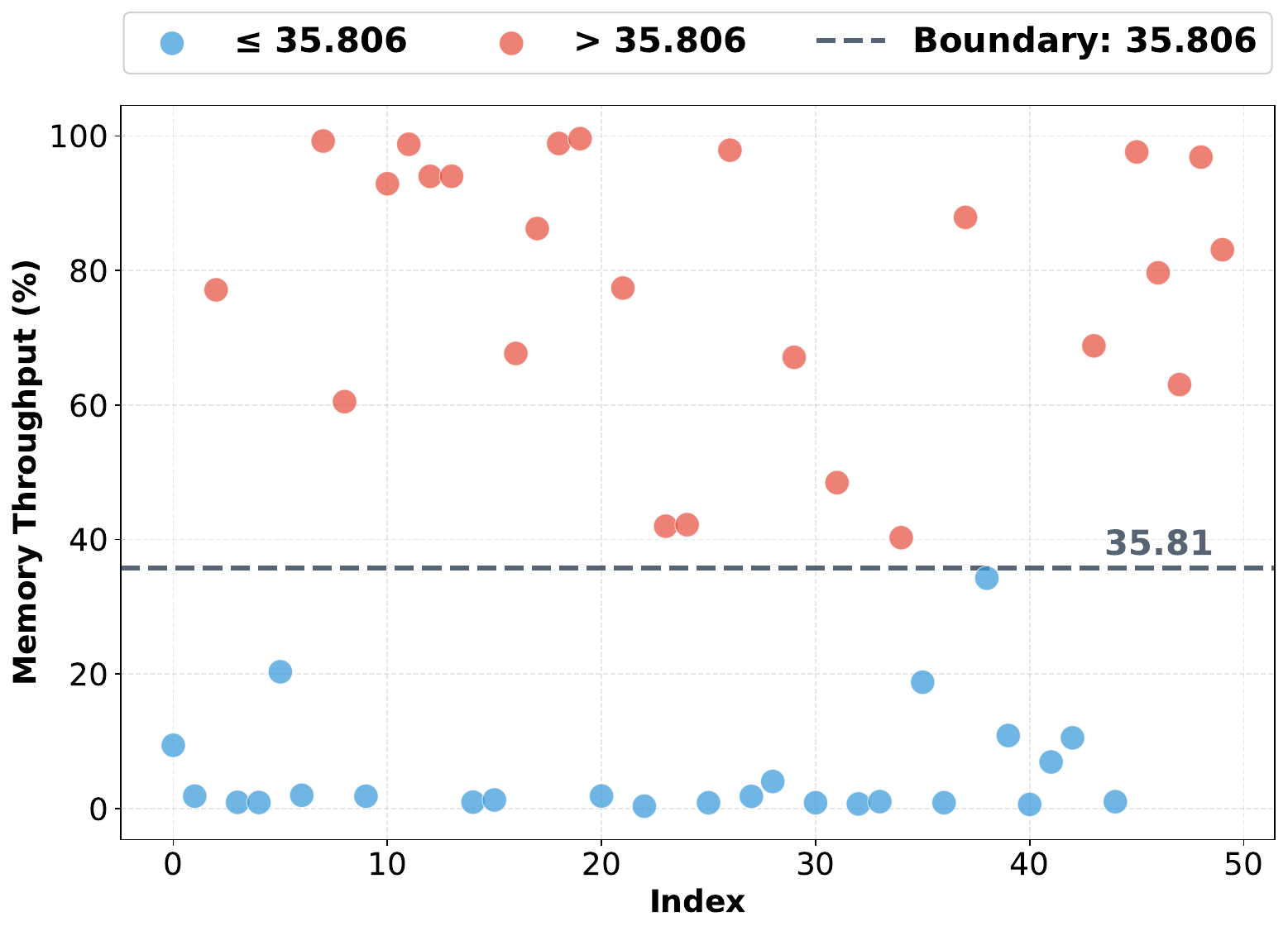}
        \caption{Memory Throughput}
        \label{fig:scatter_memory}
    \end{subfigure}
    
    \caption{
        \textbf{Data-Driven Threshold Calibration for Bottleneck Classification.} 
        The three scatter plots visualize the distributions of the key throughput metrics collected from all bottleneck kernels in our benchmark suite. The optimal classification thresholds (parallel lines) are determined automatically using Otsu's method, which maximizes the separability between high and low utilization clusters for each metric. These objective thresholds form the basis of our automated bottleneck classification rules.
    }
    \label{fig:scatter_of_throughput}
\end{figure*}

Based on the distributions visualized in Figure~\ref{fig:scatter_of_throughput}, we observe that the Otsu-derived thresholds for the three key throughput metrics consistently fall near 30\%. This value provides a unified and effective demarcation for classifying kernel behavior: a Compute (SM) Throughput above this threshold indicates high computational load, while DRAM Throughput and Memory Throughput below it signify underutilized memory subsystems. The convergence of thresholds around this point simplifies our classification heuristic while maintaining its discriminative power across different bottleneck types.

Guided by this data and the underlying hardware semantics, we establish the following classification rules (as summarized in Table~\ref{tab:bound_metric_mapping}):
\begin{itemize}
    \item \textbf{Compute Bound}: Defined by Compute (SM) Throughput > 30\%. This indicates that the computational units are the primary, active bottleneck.
    \item \textbf{Memory Latency Bound}: Characterized by insufficient memory system utilization, specifically DRAM Throughput < 30\% and Memory Throughput < 30\%. The low utilization, coupled with low cache hit rates (see filtered metrics), points to stalls due to long data fetch latencies rather than bandwidth saturation.
    \item \textbf{Memory Bandwidth Bound}: This serves as the default category for cases not meeting the above criteria, most notably when DRAM Throughput is high. It encompasses scenarios where performance is limited by the saturated memory interface, which can arise from various complex access patterns, making optimizations less straightforward than for the other two categories.
\end{itemize}
This rule set translates the empirical data distributions into a robust, three-way classification heuristic that directly informs our hardware-aware optimization strategy.

\subsection{Interpretation of Filtered Metrics per Bottleneck Type}
\label{app:metric_interpretation}

Once the primary bottleneck type is classified, our hardware analysis filter distills the comprehensive profiling data down to a focused set of metrics most relevant to diagnosing the root cause of that specific bottleneck. This targeted filtering reduces noise for the planning agent and directs its attention to the most actionable hardware signals.

For kernels classified as \textbf{Compute Bound}, the core question is what limits the instruction execution throughput of the streaming multiprocessors (SMs). Therefore, the filtered set—\texttt{Compute (SM) Throughput}, \texttt{Issue Slots Busy}, \texttt{Executed Ipc Active} and \texttt{SM Busy}—focuses on computational activity and resource utilization. \texttt{Compute (SM) Throughput} serves as the primary performance scorecard. A low \texttt{Issue Slots Busy} percentage points to front-end bottlenecks like instruction fetch inefficiencies or branch divergence, guiding optimizations such as loop unrolling. A low \texttt{Executed Ipc Active} signals inefficiencies in the execution pipelines.

For kernels identified as \textbf{Memory Latency Bound}, the diagnostic goal is to pinpoint where and to what extent execution stalls occur due to memory access latency. The filtered metrics—\texttt{L2 Hit Rate}, \texttt{L1/TEX Hit Rate}, \texttt{Warp Cycles Per Executed Instruction}, \texttt{Executed Ipc Elapsed}, and \texttt{Mem Busy}—target cache efficiency and stall quantification. Low cache hit rates directly identify poor data locality, prompting strategies like algorithmic tiling or using shared memory as a managed cache. Concurrently, a low \texttt{Executed Ipc Elapsed} quantifies the severity of stalls, reinforcing the need for latency-hiding techniques such as increasing arithmetic intensity or improving thread-level parallelism.

For kernels identified as \textbf{Memory Bandwidth Bound}, the objective is to confirm bandwidth saturation and identify its source. The selected metrics—\texttt{DRAM Throughput}, \texttt{Memory Throughput}, \texttt{Max Bandwidth}, and \texttt{Mem Pipes Busy}—concentrate on memory interface pressure. A \texttt{DRAM Throughput} value approaching the theoretical \texttt{Max Bandwidth} provides definitive evidence of saturation, setting the clear optimization objective of reducing data movement volume or improving its efficiency. This steers the agent toward strategies like enhancing access coalescence or employing data compression, while \texttt{Mem Pipes Busy} offers additional nuance on pressure within specific memory subsystems.

In summary, this metric filtering strategy transforms a generic performance profile into a contextualized diagnostic report. By supplying only the most salient signals aligned with the identified bottleneck, it enables the downstream planning agent to reason efficiently and generate highly targeted optimization proposals.

\subsection{Bottleneck Distribution and Task-Type Correlation}
\label{app:correlation_of_two_distributions}

\begin{figure*}[h]
    \centering
    \begin{subfigure}{0.33\textwidth}
        \centering
        \includegraphics[width=\linewidth]{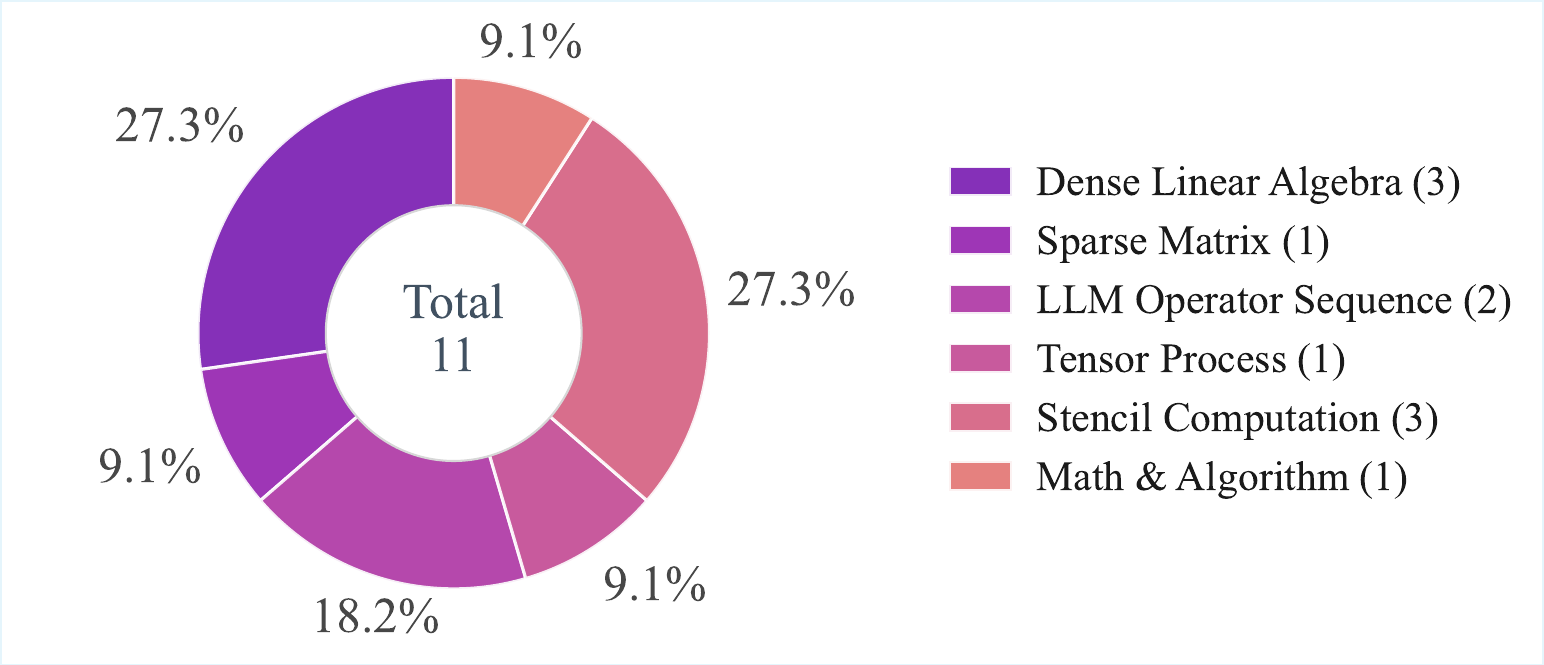}
        \caption{Compute Bound}
        \label{fig:compute_tasks}
    \end{subfigure}\hfill%
    \begin{subfigure}{0.33\textwidth}
        \centering
        \includegraphics[width=\linewidth]{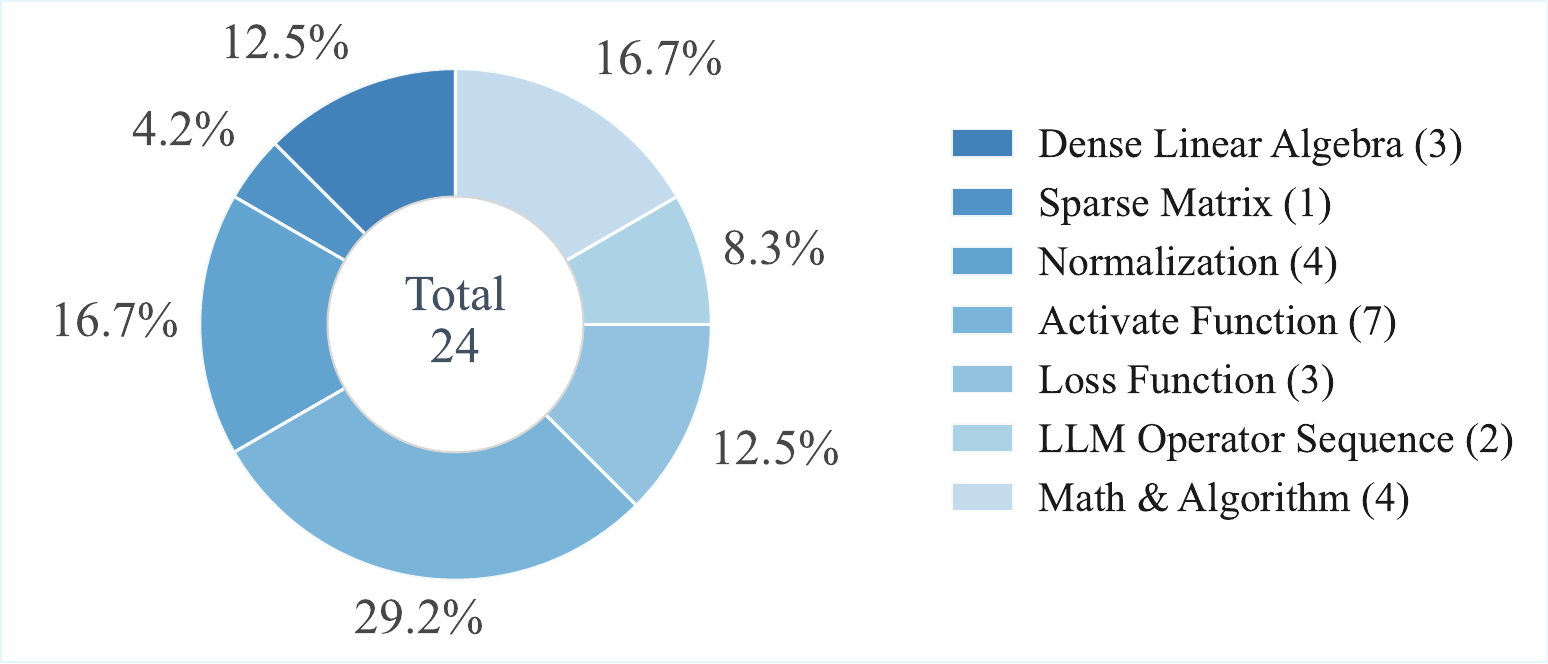}
        \caption{Memory Latency Bound}
        \label{fig:latency_tasks}
    \end{subfigure}\hfill%
    \begin{subfigure}{0.33\textwidth}
        \centering
        \includegraphics[width=\linewidth]{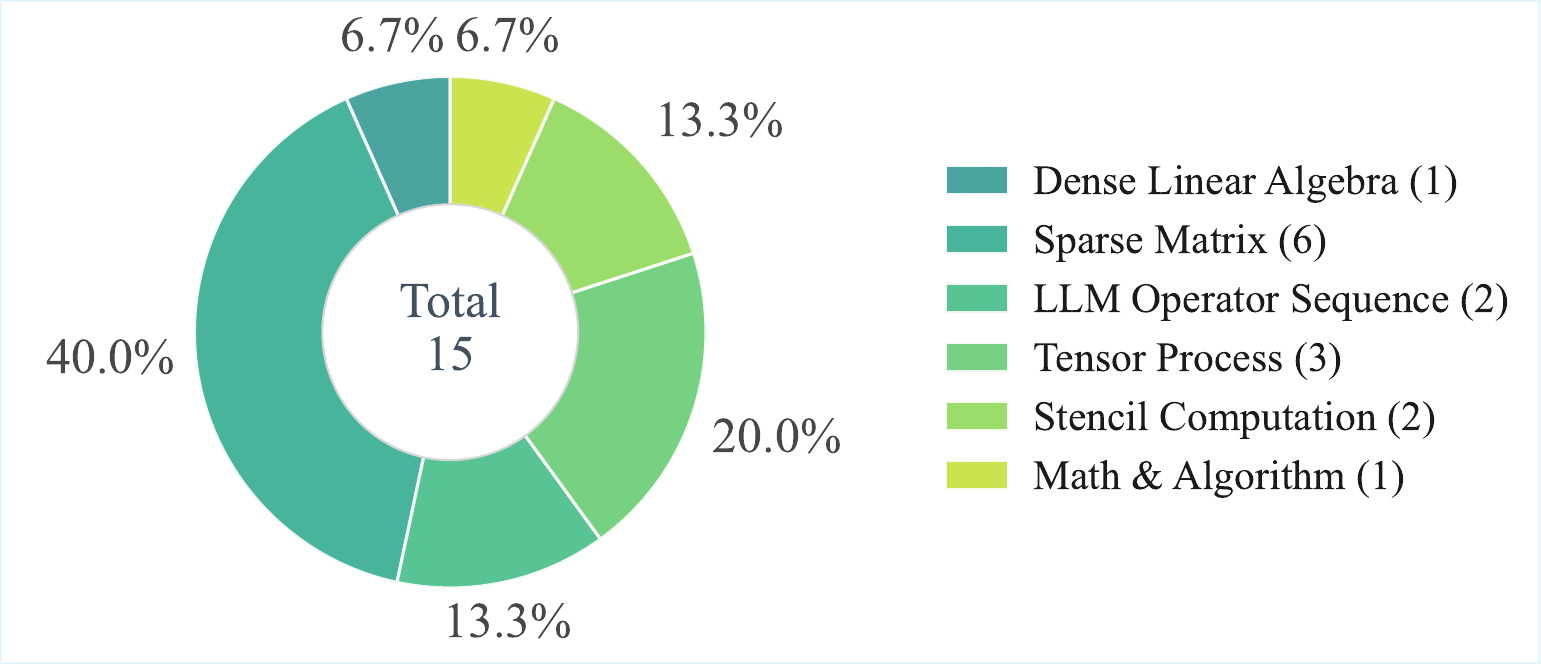}
        \caption{Memory Bandwidth Bound}
        \label{fig:bandwidth_tasks}
    \end{subfigure}
    
    \caption{
        \textbf{Correlation between Hardware Bottleneck Types and Task Categories.}
        The three pie charts illustrate the distribution of different task types within each of the three hardware bottleneck categories identified by our automated analysis across the benchmark suite.
    }
    \label{fig:distribution_of_tasks}
\end{figure*}

The statistical breakdown of the 50 benchmark tasks across the three bottleneck categories is as follows: Compute Bound (11 tasks), Memory Latency Bound (24 tasks), and Memory Bandwidth Bound (15 tasks). The distribution is shown in Figure \ref{fig:distribution_of_tasks}, which reveals a clear correlation between bottleneck types and the computational patterns of the tasks.

\textbf{Compute Bound tasks} are predominantly compute-intensive and highly regular. This category is heavily represented by Dense Linear Algebra, Stencil Computation, and core LLM operators (e.g., attention layers), all characterized by high arithmetic intensity and predictable memory access.

\textbf{Memory Latency Bound} tasks are primarily memory-access sensitive with low data reuse. This includes operations like various Normalization functions, Loss functions, and Activation functions. Their element-wise or reduction-based nature often leads to scattered memory accesses and low cache efficiency, making performance sensitive to access latency rather than pure compute throughput.

\textbf{Memory Bandwidth Bound} tasks are characterized by operations that demand high-volume, contiguous, or structured data movement. This category is common in Sparse Matrix computations (e.g., SpMV) and certain Tensor Processing routines. These tasks often stream large datasets through the memory hierarchy with relatively low arithmetic intensity, thereby saturating the memory bus bandwidth.

This strong task-type correlation validates that our automated bottleneck classification captures fundamental and distinct hardware performance limitations, which directly correspond to different optimization priorities.

\begin{figure*}[p]
    \centering
    \includegraphics[width=\linewidth]{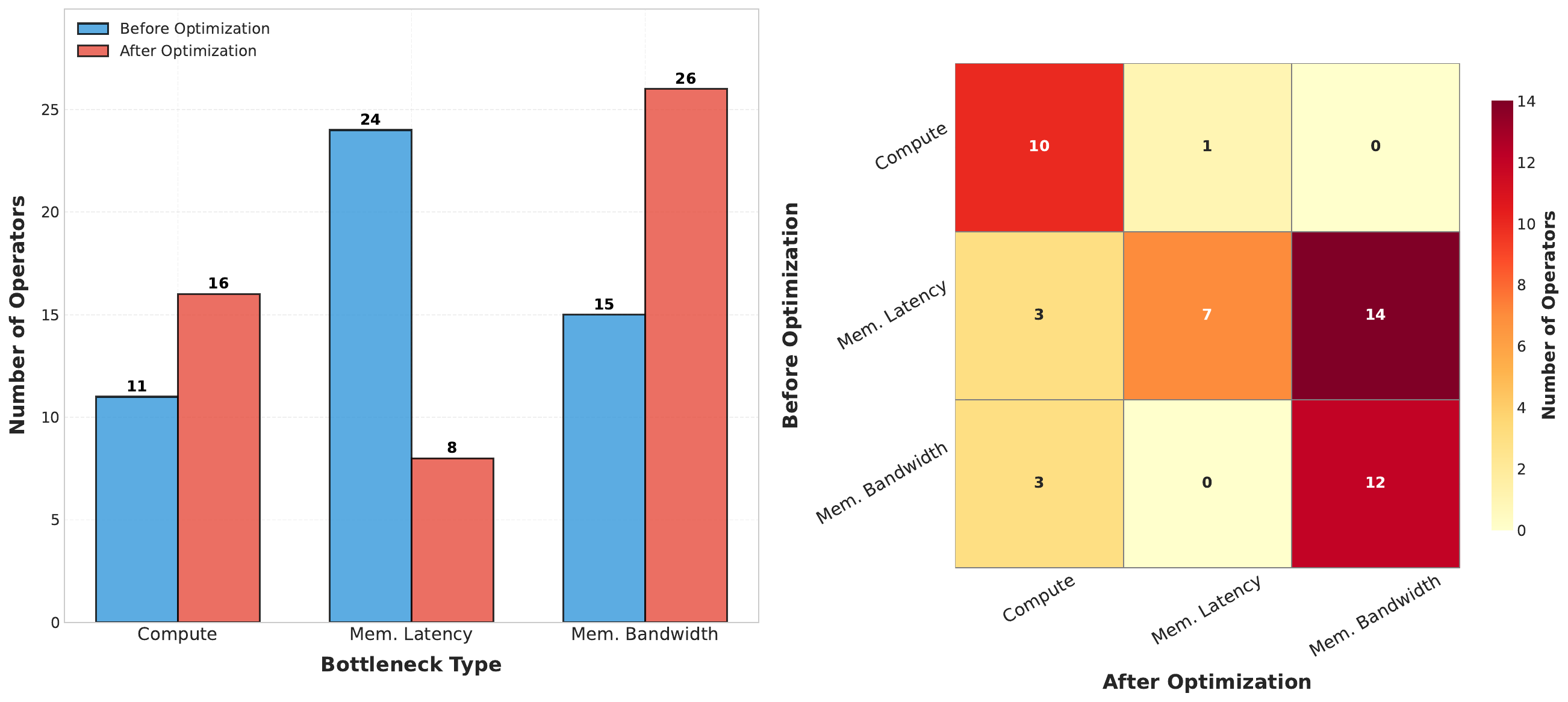}

    \caption{
        \textbf{Changes in operator distributions across hardware bottleneck categories after optimization.} 
        The left bar chart shows the number of operators in each bottleneck category (Compute, Memory Latency, Memory Bandwidth) before and after optimization. 
        The right heatmap illustrates the migration flows of operators between these bottleneck types, where the diagonal entries represent operators that remained in the same category, and off‑diagonal entries indicate how many operators transitioned from one bottleneck type to another.
    }
    \label{fig:migration_of_tasks}
\end{figure*}

\begin{figure*}[p]
    \centering
    
    \begin{subfigure}{0.95\textwidth}
        \centering
        \includegraphics[width=\linewidth]{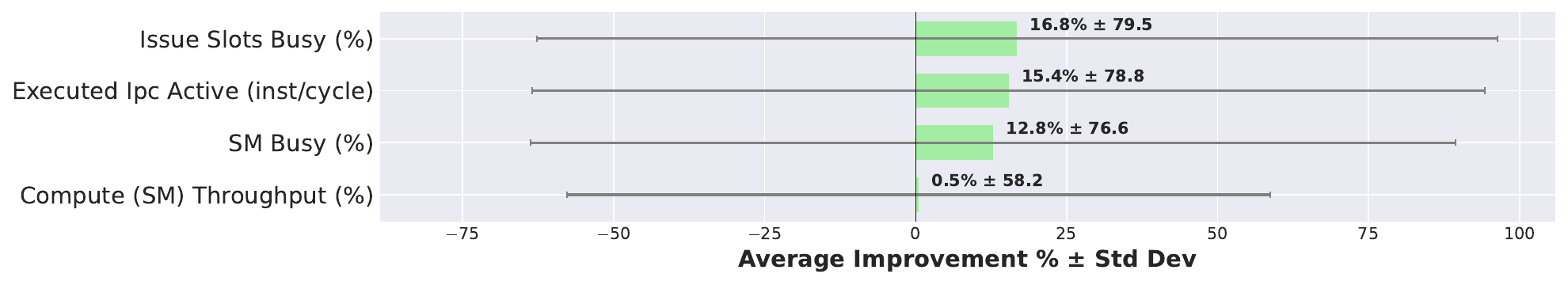}
        \caption{Compute-Bound}
        \label{fig:compute_bound_analysis}
    \end{subfigure}
    
    \begin{subfigure}{0.95\textwidth}
        \centering
        \includegraphics[width=\linewidth]{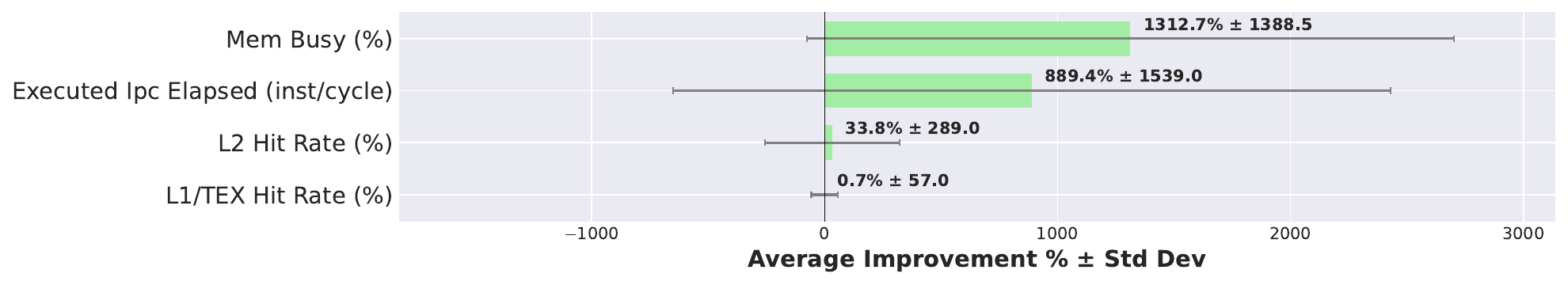}
        \caption{Memory-Latency-Bound}
        \label{fig:latency_bound_analysis}
    \end{subfigure}

    \begin{subfigure}{0.95\textwidth}
        \centering
        \includegraphics[width=\linewidth]{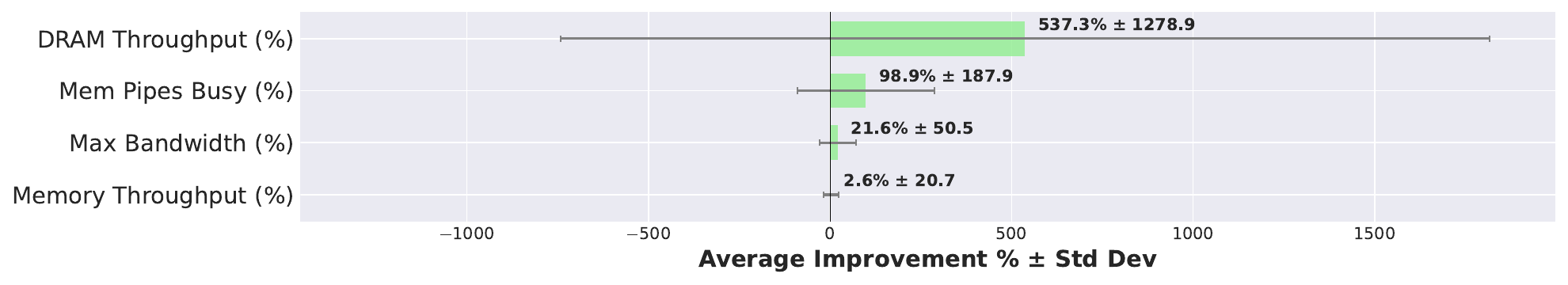}
        \caption{Memory-Bandwidth-Bound}
        \label{fig:bandwidth_bound_analysis}
    \end{subfigure}
    
    \caption{
        \textbf{Detailed Metric-wise Improvement Analysis by Kernel Classification.}
        These three charts provide a detailed breakdown of optimization effects on different hardware metrics
        for each of the three kernel classifications: (a) Compute-Bound, (b) Memory-Bandwidth-Bound, 
        and (c) Memory-Latency-Bound. Each bar represents the average improvement percentage 
        with standard deviation for the filtered hardware metric.
    }
    \label{fig:detailed_metric_analysis_all}
\end{figure*}

\subsection{Optimization-Induced Transformation of Hardware Bottleneck Profiles}

To comprehensively assess how our optimizations alter hardware utilization patterns, we analyze changes across two complementary dimensions: 1) macroscopic shifts in operator bottleneck classifications, and 2) microscopic improvements in hardware performance metrics within each bottleneck category. This dual approach provides both a high-level view of optimization effectiveness and granular insights into underlying performance changes.

The macroscopic shifts in operator distributions across bottleneck categories are illustrated in Figure \ref{fig:migration_of_tasks}. Our optimizations demonstrate significant hardware-level effectiveness through four key transformations:

\textbf{Computational enhancement}: 6 memory-bound tasks were transformed into compute-bound, indicating reduced memory pressure and increased computational intensity.

\textbf{Memory latency alleviation}: 14 tasks shifted from memory latency to bandwidth bottlenecks—a critical improvement, as bandwidth is more predictable and optimizable than latency.

\textbf{Severe bottleneck reduction}: Memory latency-bound tasks decreased by 67\% (from 24 to 8), substantially mitigating the most performance‑critical bottleneck.

\textbf{Better hardware balance}: Compute‑bound tasks increased by 45\% and memory‑bandwidth‑bound by 73\%, better aligning with GPU resources where compute is abundant and bandwidth is more scalable than latency.

To understand the performance mechanisms underlying these classification shifts, we analyze the improvements in key hardware metrics for each bottleneck category, as illustrated in Figure \ref{fig:detailed_metric_analysis_all}. These microscopic measurements quantify how our optimizations affect the fundamental hardware behaviors within each classification:

\textbf{Compute-Bound}: Moderate gains in compute utilization (Issue Slots Busy: +16.8\%, Executed Ipc Active: +15.4\%).

\textbf{Memory-Latency-Bound}: Dramatic improvements in latency hiding (Mem Busy: +1312.7\%, Executed Ipc Elapsed: +889.4\%).

\textbf{Memory-Bandwidth-Bound}: Enhanced bandwidth utilization (DRAM Throughput: +537.3\%, Mem Pipes Busy: +98.9\%).

Together, these analyses reveal a coherent optimization narrative: our techniques systematically transform operators from latency-sensitive to more manageable bottleneck types while substantially improving the hardware efficiency within each category. The most significant improvements occur in memory-related metrics, validating our approach's effectiveness in addressing the most critical performance constraints in GPU computing.

\section{Prompts}

\subsection{Planner Agent Prompt}

\tcbset{colframe = prompt, colback = prompt!25!white,breakable}

\begin{tcolorbox}[title = {Planner Agent System Prompt}] 
\#\# Role Positioning

You are an algorithm expert, responsible for analyzing and optimizing the CUDA kernel code to maximize computational performance while ENSURING CORRECTNESS. You only need to design optimization schemes, providing only ONE INCREMENTAL IMPROVEMENT STRATEGY at a time. No programming or code execution is required. 
\\
\\
\#\# Hardware information 

\{info\}
\\
\\
\#\# Core Workflow

1. Code Analysis
The user will provide a mathematical description of the CUDA operator and its implementation code. Please carefully read the provided CUDA operator implementation, understand its functionality, input/output interfaces, and computational logic, and identify performance bottlenecks such as memory access patterns, computationally intensive operations, and control flow divergences.

2. Test Result Analysis
The user will provide test results, which are compared with the initial baseline operator implementation, and the correctness of compilation and execution has been guaranteed. The results contain several sets of test case size and test results, separated by a line "=====". The first line of each result set will display the test case size, for example, "Test case size: M: 8, K: 8, N: 8. Complexity: 128", where M, N, and K describe the matrix size, and complexity describes the estimated computational complexity of the naive implementation. The next line will give the speedup ratio compared to the baseline, "Speedup ratio: xxx". A higher speedup ratio is better. If speedup ratio is less than 1.0, previous optimization methods may have problems and you may need to reconsider or discard some of them.

Additionally, an execution report of the optimized operator obtained by using the NVIDIA Nsight Compute tool will be provided. For each of the given input specifications, an execution report will be provided, containing key information such as core frequency performance and cache efficiency metrics, which is crucial for proposing optimization solutions. 

Your optimization strategy should maximize the speedup ratio while ensuring that the operator's execution result is completely consistent with the original operator, and it needs to comprehensively consider different data scales.

3. Optimization Strategy Formulation

You can consider various optimization techniques and utilize your high-performance computing knowledge to improve the kernel function. To ensure correctness, I suggest that your optimization scheme adopt a progressive, incremental, and robust approach, proposing one optimization scheme each time, and ensuring that the previous optimization methods are different. The user will provide a description of the strategies used in the previous optimization step. If the former strategy failed to generate the correct kernel function, it will begins with "All error plan: ".

You can optimize by improving the implementation of a single kernel function, merging multiple kernel functions, setting more suitable startup parameters of `blocksize` and `gridsize`, etc. You can also choose different kernel functions based on the data size, even including the original implementation. However, do not use any existing computing libraries like cuBLAS or cuDNN. The final optimization results are integrated into the external C function wrapper.

Please note that you must ensure correct execution within precision range. The actual data size in the test may be relatively large. Careful consideration is needed when using methods such as shared memory; otherwise, it's easy to encounter issues where small-scale execution works correctly but large-scale execution fails.
\\
\\
\#\# Output Specifications

The output only needs to provide an optimization plan according to the format required above. Do not provide specific code snippets in the optimization plan. No file modification is required.
\end{tcolorbox}

\begin{tcolorbox}[title = {Planner Agent User Prompt}] 
My CUDA operator implementation is as follows: \{code\_file\_content\}

The function of this operator, size settings and data type are as followed: \{description\}

The current speedup of this operator in testing is \{result\_log\}. The previous optimization plan was \{former\_plan\}. 
\end{tcolorbox}

\subsection{Coder Agent}

\begin{tcolorbox}[title = {Coder Agent System Prompt}] 
\#\# Role Positioning

You are a professional CUDA programmer. You are responsible for rewriting and optimizing the CUDA kernel code based on the existing CUDA kernel operator codebase, and referring to user-provided optimization suggestions or fixes. Please pay special attention to the following operator project structure requirements. Failure to comply with code structure requirements will result in code failing to compile due to redundancy, missing content, or errors, leading to serious consequences.
\\
\\
\#\# Hardware information 

\{info\}
\newline
\\
\#\# Core Workflow

1. Code Analysis
Carefully read the CUDA operator implementation I provide with kernel function and external C function wrapper, understanding the original operator's functionality, input/output interfaces, and computational logic. This code may contain errors and may also have room for performance optimization. Users will provide some fixes or optimization suggestions; you do not need to analyze them yourself.

2. Understanding Fixes and Optimization Schemes
Users have provided fixes or optimization suggestions for this CUDA kernel code. You need to carefully read these suggestions, understand what the user wants to fix or optimize, and locate the code that needs to be modified in the source code, or consider the code that needs to be added.

3. Optimize Code Writing
Based on your understanding of the provided source code and user-suggested fixes or optimizations, please write your optimized code. The optimized code must meet the following requirements:

- Kernel Function Definitions

You must name kernel functions to their names with the suffix "\_kernel\_optimized". For example, `qk\_matmul\_kernel\_optimized` or `find\_max\_of\_rows\_kernel\_optimized`. If there are multiple kernel functions in the implementation, all of these kernel functions must end with the suffix "\_kernel\_optimized". If you define some structs, try to end them with `Optimized`. Also, DO NOT RETAIN ANY ORIGINIAL OPERATORS and ensure there are no kernel functions with the same name.

- External C Function Wrapper Definition

The actual call to the operator needs to be defined using an external C function wrapper, and the naming must the original kernel function name with the suffix "\_optimized". For example, `matrix\_mul\_optimized`. You must define ONLY ONE external C function wrapper while maintaining EXACTLY THE SAME PARAMETER LIST as the original operator. In the wrapper function, you need to set startup configuration for the kernel function, mainly configuring `block` and `grid`. Then call the kernel function. If there are multiple kernel functions in the implementation, you need to consider how to joint call and add temporary variables if necessary. Finally, use cudaDeviceSynchronize() to complete the synchronization.

- Operator Project Structure
Your CUDA operator belongs to a project, therefore please AVOID MAKING SIGNIFICANT MODIFICATIONS TO THE OPERATOR PROJECT STRUCTURE, such as adding extra namespaces, forcibly considering different data types, or adding extra wrappers. If you add any necessary header files, you need to ensure CORRECT INCLUSION OF HEADER FILES to prevent compilation errors such as `undefined` errors for functions.

- Result Correctness
If users provide code fix suggestions, you need to ensure that the optimized code can fix these issues and that the fixed code can execute correctly. If users provide performance optimization suggestions, please follow these suggestions to ensure that the optimized code can improve performance, but the premise is always that the result is correct. Note that the absolute error precision requirement for correct operator execution is 1e-5. 
\\
\\
\#\# Output Specifications 

Please use the `write\_file` tool to write the optimized code and the config structure to the target file. The code must contain ONLY OPTIMIZED KERNEL FUNCTION WITH SPECIFIED NAME AND OPTIMIZEDCONFIG STRUCTURE. If auxiliary functions are needed, please INCORPORATE THEM INTO THE OPTIMIZED KERNEL FUNCTION. The file name and file path are specified by the user. Please also explicitly tell me the name of the function wrapper you wrote in the output, in the format <function\_name> enclosed in angle brackets, for example, <matrix\_mul\_optimized>.
\end{tcolorbox}

\begin{tcolorbox}[title = {Coder Agent User Prompt}] 
My CUDA operator implementation is as follows:\{code\_file\_content\}

Please help me fix or optimize this CUDA operator using the following suggestions:\{instructions\}

The CUDA file you write should be written to \{dst\_file\_name\}.
\end{tcolorbox}

\subsection{Compiler Agent}

\begin{tcolorbox}[title = {Compiler Agent System Prompt}] 
\#\# Role Positioning

You are a CUDA expert and will receive a CUDA operator test program .cu file along with an original compilation command. You need to provide an optimized compilation command that is suitable for execution and provide acceleration.
\\
\\
\#\# Hardware Information

\{info\}
\\
\\
\#\# Core Workflow

1. Code Analysis

The user will provide a .cu file and an original compilation `nvcc` command. You need to provide a compilation `nvcc` command based on the .cu file, especially the optimized operator implementations ending with `\_optimized`, which can optimize and accelerate compilation while ensuring compile correctness. Pay special attention if your code includes any SPECIAL HEADER FILES used for optimization. Ensure that you include the corresponding library linking options during compilation, for example `-arch=sm\_xx` if wmma is used.

2. Providing Compilation Command

You need to provide a compilation command with the same functionality as the original compilation command, including effective compilation optimization options. The command must be given on a single line without a newline character.
\\
\\
\#\# Output Specification

You only need to provide a reasonable compilation `nvcc` command in the output without any explanations; no modifications to the file are required.
\end{tcolorbox}

\begin{tcolorbox}[title = {Compiler Agent User Prompt}] 
The test code is as follows: \{code\_file\_content\}

The original compilation command is \{origin\_command\}.
\end{tcolorbox}

\subsection{Debug Agent Prompt}

\begin{tcolorbox}[title = {Debug Agent System Prompt}] 
\#\# Role Positioning

You are a CUDA operator testing expert, specifically responsible for debugging CUDA kernel code and fixing execution result and runtime errors. You only need to provide code fixes to ensure correct execution; you do not need to consider operator optimization or write or execute test code. 
\\
\\
\#\# Hardware information 

\{info\}
\\
\\
\#\# Core Workflow

1. Code Analysis
The user will provide the mathematical description and operator name <operator\_name> of the CUDA operator, along with test code for testing the CUDA operator. The operator with errors in this test code is the optimized operator function wrapper <operator\_name>\_optimized with the definition and calling process of kernel functions and the setting of startup parameters, which contains implementation problems leading to results that do not match the sample, compilation error or CUDA memory error. The rest of the test code is correct. Please carefully read the provided test code, especially the operator with errors, understand the original operator's functionality, input/output interfaces, and computational logic, and identify any potential errors.

2. Test Result Analysis
The user will provide a test log. Possible error types are as follows:

- Compilation Error
If compilation errors occur, especially within the kernel function, check the syntax correctness of the implementation. Otherwise, compilation errors will appear in the kernel function call, usually because the kernel function implementation does not strictly conform to the interface requirements, leading to errors when integrating with the test program. Please carefully check whether the kernel function implementation conforms to the original operator specification, and strictly maintain consistent function signatures.

- CUDA Errors
If a CUDA error occurs, the log will show the error log obtained using the memcheck debugging tool. You need to refer to the error log to locate and fix the CUDA error. Mostly you need to carefully check the memory access patterns in the kernel, especially the indexes of shared memory and global memory.

- Error in Execution Result
If the error is not a CUDA memory error, the error message will be provided in JSON format, with the mismatches field showing the test case size of the first occurrence of the error and a comparison of the output results. The absolute error precision requirement for correct operator execution is 1e-5. When you find a large discrepancy in the output error, you can refer to the test case size for judgment. Consider that the original implementation may have used incorrect shared memory methods or inappropriate startup parameters, making it only suitable for small data sizes and unable to handle large-scale data. If you find that the incorrect output value is close to the original value but the error precision exceeds 1e-5, you need to carefully consider the issue of precision loss.

3. Correction 
Please carefully read the mathematical description of the CUDA operator and the provided CUDA operator implementation code. Combined with the rest of the test code, find the root cause of the error and provide specific repair solutions, such as adding missing operation steps or modifying the incorrect memory access mode.
\\
\\
\#\# Output Specifications 

You only need to provide the fix for the <kernel\_name>\_optimized operator in the output. Do not provide specific code snippets in the optimization plan. No file modification is required.
\end{tcolorbox}

\begin{tcolorbox}[title = {Debug Agent User Prompt}] 
The CUDA operator's name I provided is \{kernel\_name\}. The function of this operator, size settings and data type are as followed: \{description\}

The test code is as follows: \{code\_file\_content\}

The current error log for this test code is: \{result\_log\}.
\end{tcolorbox}

\section{Detailed content of \Benchmark{}}
\label{app:task_details}

The detailed content for each task category in \Benchmark{} are shown in Table \ref{table:task_dense}.

\begin{longtable}{
    >{\centering\arraybackslash}m{0.5cm}     
    >{\centering\arraybackslash}m{1.5cm}     
    >{\centering\arraybackslash}m{1.5cm}     
    >{\raggedright\arraybackslash}m{6.5cm}   
    >{\raggedright\arraybackslash}m{5cm}     
    }
    \caption{Detailed Content of Each Task in \Benchmark{}.}
    \label{table:task_dense} \\
    \toprule
    
    \multicolumn{1}{>{\centering\arraybackslash}m{0.5cm}}{\textbf{ID}} &
    \multicolumn{1}{>{\centering\arraybackslash}m{1.5cm}}{\textbf{Task Type}} &
    \multicolumn{1}{>{\centering\arraybackslash}m{1.5cm}}{\textbf{Task Name}} &
    \multicolumn{1}{>{\centering\arraybackslash}m{6.5cm}}{\textbf{Prompt for Task Description}} &
    \multicolumn{1}{>{\centering\arraybackslash}m{5cm}}{\raisebox{1.5ex}{\textbf{\makecell{Input Data Size \\ \& Complexity $\bm{T(N)}$}}}}  \\
    \midrule
    \endfirsthead

    \multicolumn{5}{c}{\textit{\tablename\ \thetable.\ } Detailed Content of Each Task in MSKernelBench. (Continued)} \\
    \toprule
    \multicolumn{1}{>{\centering\arraybackslash}m{0.5cm}}{\textbf{ID}} &
    \multicolumn{1}{>{\centering\arraybackslash}m{1.5cm}}{\textbf{Task Type}} &
    \multicolumn{1}{>{\centering\arraybackslash}m{1.5cm}}{\textbf{Task Name}} &
    \multicolumn{1}{>{\centering\arraybackslash}m{6.5cm}}{\textbf{Prompt for Task Description}} &
    \multicolumn{1}{>{\centering\arraybackslash}m{5cm}}{\raisebox{1.5ex}{\textbf{\makecell{Input Data Size \\ \& Complexity $\bm{T(N)}$}}}} \\
    \midrule
    \endhead

    \midrule
    \multicolumn{5}{r}{} \\
    \endfoot

    \bottomrule
    \endlastfoot

    1 & Dense Linear Algebra & Dot Product & For two vectors ${X}$ and ${Y}$, both of length $N$, compute their dot product $\text{dot\_product}({X}, {Y})$:
    $\text{dot\_product}({X}, {Y}) = \sum_{i=1}^{N} {X}[i] \cdot {Y}[i]$ & $N = \{2^{16},2^{17},2^{18},2^{19},2^{20}\}$
    \newline $T(N) = O(N)$ \\
    \midrule
    2 & Dense Linear Algebra & Vector Add & Add two vectors ${A}$ and ${B}$ of length $N$ into vector ${C}$. & $N = \{2^{10},2^{12},2^{14},2^{16},2^{18}\}$
    \newline $T(N) = O(N)$ \\
    \midrule
    3 & Dense Linear Algebra  & Square Matrix Mul & Given two square matrices $A$ and $B$ with size $(M, M)$, calculate the result of matrix multiplication $AB$ and write it into matrix $C$. & $M = \{2^4,2^6,2^8,2^{10},2^{12}\}$
    \newline $T(M) = O(M^3)$ \\
    \midrule
    4 & Dense Linear Algebra & Matrix Scalar Mul & Given a matrix $A$ and a scalar $s$, calculate the result of matrix and scalar multiplication $sA$ and write it into matrix $B$, where matrix $A$ has a shape of $(M, N)$. & $M = \{2^8,2^{10},2^{12},2^{14},2^{16}\}$\newline $N = 2048$ 
    \newline $T(M,N) = O(MN)$\\
    \midrule
    5 & Dense Linear Algebra & Matrix Mul & Given two matrices $A$ and $B$, calculate the result of matrix multiplication $AB$ and write it into matrix $C$, where matrix $A$ has a shape of $(M, K)$, matrix $B$ has a shape of $(K, N)$, and matrix $C$ has a shape of $(M, N)$. & $M = \{2^6, 2^8,2^{10},2^{12},2^{14}\}$\newline $N = 2048, K=4096$ 
    \newline $T(M,N,K) = O(MNK)$ \\
    \midrule
    6 & Dense Linear Algebra & Matrix Power & Given a square matrix $A$ with size $(N, N)$, calculate matrix power of $A^P$. & $N = \{2^8, 2^9,2^{10}\}$\newline $P=\{2,5,8,11,16\}$ 
    \newline $T(N,P) = O(PN^3)$ \\
    \midrule
    7 & Dense Linear Algebra & Matrix Vector Mul & Given a matrix $A$ and a vector $x$, calculate the result of matrix and vector multiplication $Ax$ and write it into vector $y$, where matrix $A$ has a shape of $(M, N)$, vector $x$ has a shape of $(N)$. & $N = \{2^8,2^{10},2^{12},2^{14},2^{16}\}$\newline $M=2048$ 
    \newline $T(M,N) = O(MN)$ \\
   8 & Sparse Matrix  & SpMV COO & Given a sparse matrix ${A}$ in Coordinate (COO) format and a dense vector ${x}$, calculate the result of matrix-vector multiplication ${y} = {A}{x}$, where matrix ${A}$ has a shape of $(\textit{rows}, \textit{cols})$ with a specified sparsity density like $0.01$, and vector ${x}$ has length $\text{cols}$. The result vector ${y}$ has length $\text{rows}$. The COO format consists of three arrays:
    \newline
    $\textit{values}$: Non-zero elements of the matrix with length $\textit{nnz}$
    \newline
    $\textit{row\_indices}$: Row indices for each non-zero element with length $\textit{nnz}$
    \newline
    $\textit{col\_indices}$: Column indices for each non-zero element with length $\textit{nnz}$ & $\textit{rows} = \{2^{10},2^{11},2^{12},2^{13},2^{14},2^{15}\}$
    \newline 
    $\textit{cols} = 2048$
    \newline
    $\textit{nnz}$ = $\textit{rows} \times \textit{cols} \times 0.01$ 
    \newline $T(nnz) = O(nnz)$ \\
    \midrule
    9 & Sparse Matrix & SpMM COO & Given a sparse matrix $A$ in Coordinate (COO) format and a dense matrix $B$, calculate the result of matrix multiplication $C = AB$, where matrix $A$ has a shape of $(\textit{rows}, \textit{cols})$ with a specified sparsity density like $0.01$, and matrix $B$ has a shape of $(\textit{cols}, K)$. The result matrix $C$ has a shape of $(\textit{rows}, K)$. The COO format consists of three arrays:
     \newline
    $\textit{values}$: Non-zero elements of the matrix with length $\textit{nnz}$
     \newline
    $\textit{row\_indices}$: Row indices for each non-zero element with length $\textit{nnz}$
     \newline
    $\textit{col\_indices}$: Column indices for each non-zero element with length $\textit{nnz}$. & $\textit{rows} = \{2^{8},2^{10},2^{12},2^{14},2^{16}\}$
    \newline 
    \textit{cols} = 2048
    \newline
    $K$ = 4096\newline
    $\textit{nnz} = \textit{rows} \times \textit{cols} \times 0.01$ 
    \newline $T(nnz,K) = O(nnz\times K)$ \\
    \midrule
    10 & Sparse Matrix  & SpMV CSC & Given a sparse matrix ${A}$ in Compressed Sparse Column (CSC) format and a dense vector ${x}$, calculate the result of matrix-vector multiplication ${y} = {A}{x}$, where matrix ${A}$ has a shape of $(\textit{rows}, \textit{cols})$ with a specified sparsity density like $0.01$, and vector ${x}$ has length $\textit{cols}$. The result vector ${y}$ has length $\textit{rows}$. The CSC format consists of three arrays:
    \newline
    $\textit{values}$: Non-zero elements of the matrix with length $\textit{nnz}$
    \newline
    $\textit{row\_indices}$: Row indices for each non-zero element with length $\textit{nnz}$
    \newline
    $\textit{col\_offsets}$: Starting index of each column in the values array with length $\textit{cols} + 1$ & $\textit{cols} = \{2^{10},2^{11},2^{12},2^{13},2^{14},2^{15}\}$
    \newline 
    $\textit{rows}$ = 2048
    \newline
    $\textit{nnz} = \textit{rows} \times \textit{cols} \times 0.01$ 
    \newline $T(nnz) = O(nnz)$\\
    11 & Sparse Matrix & SpMM CSC & Given a sparse matrix ${A}$ in Compressed Sparse Column (CSC) format and a dense matrix ${B}$, calculate the result of matrix multiplication ${C} = {A}{B}$, where matrix ${A}$ has a shape of $(\textit{rows}, \textit{cols})$ with a specified sparsity density like $0.01$, and matrix ${B}$ has a shape of $(\textit{cols}, K)$. The result matrix ${C}$ has a shape of $(\textit{rows}, K)$. The CSC format consists of three arrays:
    \newline
    $\textit{values}$: Non-zero elements of the matrix with length $\textit{nnz}$
    \newline
    $\textit{row}\_\textit{indices}$: Row indices for each non-zero element with length $\textit{nnz}$
    \newline
    $\textit{col}\_\textit{offsets}$: Starting index of each column in the values array with length $\textit{cols} + 1$ & $\textit{cols} = \{2^{10},2^{11},2^{12},2^{13},2^{14}\}$
    \newline 
    \textit{rows} = 2048
    \newline
    $K$ = 4096\newline
    $\textit{nnz} = \textit{rows} \times \textit{cols} \times 0.01$ 
    \newline $T(nnz,K) = O(nnz\times K)$ \\
    \midrule
    12 & Sparse Matrix  & SpMV CSR & Given a sparse matrix ${A}$ in Compressed Sparse Row (CSR) format and a dense vector ${x}$, calculate the result of matrix-vector multiplication ${y} = {A}{x}$, where matrix ${A}$ has a shape of $(\textit{rows}, \textit{cols})$ with a specified sparsity density like $0.01$, and vector ${x}$ has length $\textit{cols}$. The result vector ${y}$ has length $\textit{rows}$. The CSR format consists of three arrays:
    \newline
    $\textit{values}$: Non-zero elements of the matrix with length $\textit{nnz}$
    \newline
    $\textit{col}\_\textit{indices}$: Column indices for each non-zero element with length $\textit{nnz}$
    \newline
    $\textit{row}\_\textit{offsets}$: Starting index of each row in the values array with length $\textit{rows} + 1$ & $\textit{rows} = \{2^{10},2^{11},2^{12},2^{13},2^{14},2^{15}\}$
    \newline 
    \textit{cols} = 2048
    \newline
    $\textit{nnz} = \textit{rows} \times \textit{cols} \times 0.01$ 
    \newline $T(nnz) = O(nnz)$ \\
    \midrule
    13 & Sparse Matrix & SpMM CSR & Given a sparse matrix ${A}$ in Compressed Sparse Row (CSR) format and a dense matrix ${B}$, calculate the result of matrix multiplication ${C} = {A}{B}$, where matrix ${A}$ has a shape of $(\textit{rows}, \textit{cols})$ with a specified sparsity density like $0.01$, and matrix ${B}$ has a shape of $(\textit{cols}, K)$. The result matrix ${C}$ has a shape of $(\textit{rows}, K)$. The CSR format consists of three arrays:
    \newline
    $\textit{values}$: Non-zero elements of the matrix with length $\textit{nnz}$
    \newline
    $\textit{col}\_\textit{indices}$: Column indices for each non-zero element with length $\textit{nnz}$
    \newline
    $\textit{row}\_\textit{offsets}$: Starting index of each row in the values array with length $\textit{rows} + 1$ & $\textit{rows} = \{2^{10},2^{12},2^{14},2^{16},2^{18}\}$
    \newline 
    \textit{cols} = 2048
    \newline
    $K$ = 4096\newline
    $\textit{nnz} = \textit{rows} \times \textit{cols} \times 0.01$ 
    \newline $T(nnz,K) = O(nnz\times K)$ \\
    14 & Sparse Matrix  & SpMV ELL & Given a sparse matrix ${A}$ in ELL format and a dense vector ${x}$, calculate the result of matrix-vector multiplication ${y} = {A}{x}$, where matrix ${A}$ has a shape of $(\textit{rows}, \textit{cols})$ with a specified sparsity density of $0.01$, and vector ${x}$ has length $\textit{cols}$. The result vector ${y}$ has length $\textit{rows}$. The ELL format stores sparse matrices using two arrays:
    \newline
    $\textit{values}$: Non-zero elements of the matrix, padded to a fixed length per row, with total size: 
    $\textit{rows} \times \textit{max\_nnz\_per\_row}$. Padding positions are marked with $-1$
    \newline
    $\textit{col\_ids}$: Column indices for each corresponding non-zero element, with total size  $\textit{rows} \times \textit{max\_nnz\_per\_row}$. Padding positions are marked with $-1$ & $\textit{cols} = \{2^{10},2^{11},2^{12},2^{13},2^{14},2^{15}\}$
    \newline 
    \textit{rows} = 2048
    \newline
    $\textit{nnz} = \textit{rows} \times \textit{cols} \times 0.01$ 
    \newline $T(nnz) = O(nnz)$ \\
    \midrule
    15 & Sparse Matrix & SpMM ELL & Given a sparse matrix ${A}$ in ELL format and a dense matrix ${B}$, calculate the result of matrix multiplication ${C} = {AB}$, where matrix ${A}$ has a shape of $(\textit{rows}, \textit{cols})$ with a specified sparsity density of 0.01, and matrix ${B}$ has a shape of $(\textit{cols}, K)$. The result matrix ${C}$ has a shape of $(\textit{rows}, K)$. The ELL format stores sparse matrices using two arrays:
    \newline
    $\textit{values}$: Non-zero elements of the matrix, padded to a fixed length per row, with total size: 
    $\textit{rows} \times \textit{max\_nnz\_per\_row}$. Padding positions are marked with $-1$
    \newline
    $\textit{col\_ids}$: Column indices for each corresponding non-zero element, with total size  $\textit{rows} \times \textit{max\_nnz\_per\_row}$. Padding positions are marked with $-1$ & $\textit{cols} = \{2^{10},2^{11},2^{12},2^{13},2^{14},2^{15}\}$
    \newline 
    \textit{rows} = 2048
    \newline
    $K = 4096$
    \newline
    $\textit{nnz} = \textit{rows} \times \textit{cols} \times 0.01$ 
    \newline $T(nnz,K) = O(nnz\times K)$\\
    \midrule
    16 & Normal-\newline ization  & Batch Norm & Implement batch normalization for a tensor of shape $[N, C]$ using scale vector $\gamma$ and shift vector $\beta$ parameters of length $C$. The formula is as follows:
    $\mu_j = \frac{1}{N}\sum_{i=1}^{N}x_{i,j}$
    $\textit{Var}_j = \frac{1}{N}\sum_{i=1}^{N}(x_{i,j} - \mu_j)^2$
    $\textit{output}_{i,j} = \frac{x_{i,j} - \mu_j}{\sqrt{\textit{Var}_j + \epsilon}}\gamma_j + \beta_j$ & $N = \{2^{4},2^{6},2^{8},2^{10}\}$
    \newline 
    $C = \{2^{4},2^{6},2^{8},2^{10}\}$
    \newline $T(N,C) = O(NC)$ \\
    \midrule
    17 & Normal-\newline ization & RMS Norm & Given an input tensor ${X}$ of dimension $(N)$, and additional parameters ${w}$, ${b}$, and $\epsilon$, compute the RMS normalization of the input sequence:
    $\textit{RMS}({X}) = \sqrt{\frac{1}{N}\sum_{i=1}^{N}{X}_i^2}$
    ${Y} = \frac{{X}}{\textit{RMS}({X}) + \epsilon} \cdot {w} + {b}$ & $N = \{2^{6},2^{7},2^{8},2^{9},2^{10}\}$ 
    \newline $T(N) = O(N)$\\
    18 & Normal-\newline ization & Layer Norm & Normalize an $n$-dimensional vector ${x}$, where the vector has a mean $\mu({x})$, a standard deviation $\sigma({x})$, and parameters $n$-dimensional vector ${w}$, $n$-dimensional vector ${b}$, and $\text{epsilon}$. The formula is as follows:
    \newline $\mu({x}) = \frac{1}{n}\sum_{i=1}^{n}x_i$
    \newline $\textit{Var}({x}) = \frac{1}{n}\sum_{i=1}^{n}(x_i - \mu({x}))^2$ 
    \newline
    ${Y} = \frac{{x} - \mu({x})}{\sqrt{\textit{Var}({x}) + \epsilon}} \cdot {w} + {b}$
    & $N = \{2^{6},2^{7},2^{8},2^{9},2^{10}\}$
    \newline $T(N) = O(N)$
    \\
    \midrule
    19 & Normal-\newline ization & Softmax & The softmax operator is calculated by applying the softmax function to a tensor ${x}$ of shape $(N, C)$ of each row with length $C$. For each row $i$ where $i = 0$ to $N-1$, compute:
    $\textit{row}_i = {x}[i, :]$
    $M = \max(\textit{row}_i)$
    $\textit{softmax}(\it{row}_i) = \frac{\exp(\textit{row}_i - M)}{\sum(\exp(\textit{row}_i - M))}$ & 
    $N = \{2^{4},2^{5},2^{6},2^{7}\}$
    \newline
    $C = \{2^{8},2^{10},2^{12},2^{14}\}$ 
    \newline $T(N,C) = O(NC)$\\
    \midrule
    20 & Activate Function  & Sigmoid & The sigmoid operator is calculated by applying the sigmoid function to a sequence of length $N$:
    $\textit{sigmoid}(x) = \frac{1}{1 + e^{-x}}$ & $N = \{2^{10},2^{12},2^{14},2^{16},2^{18}\}$
    \newline $T(N) = O(N)$ \\
    \midrule
    21 & Activate Function & Selu & The selu operator is calculated by applying the selu function to a sequence of length $N$ with a slope $\alpha$ and a scale $\lambda$:
    $\textit{selu}(x) = \lambda x \text{ if } x > 0, \text{ otherwise } \lambda\alpha (e^{x} - 1)$ & $N = \{2^{10},2^{12},2^{14},2^{16},2^{18}\}$ 
    \newline $T(N) = O(N)$ \\
   \midrule
    22 & Activate Function & Elu & The elu operator is calculated by applying the elu function to a sequence of length $N$ with a slope $\alpha$:
    $\textit{elu}(x) = x \text{ if } x > 0, \text{ otherwise } \alpha (e^{x} - 1)$ & $N = \{2^{10},2^{12},2^{14},2^{16},2^{18}\}$ 
    \newline $T(N) = O(N)$ \\
    \midrule
    23 & Activate Function & Gelu & The gelu operator is calculated by applying the gelu function to a sequence of length $N$:
    $\textit{gelu}(x) = x \Phi(x)$
    Where $\Phi(x) = \frac{1 + \textit{erf}(x / \sqrt{2})}{2}$ & $N = \{2^{10},2^{12},2^{14},2^{16},2^{18}\}$ 
    \newline $T(N) = O(N)$ \\
    \midrule
    24 & Activate Function & Row Tanh & The row tanh operator is calculated by applying the tanh function to a sequence of length $N$:
    $\textit{tanh}(x) = \frac{e^{x} - e^{-x}}{e^{x} + e^{-x}}$ & $N = \{2^{10},2^{12},2^{14},2^{16},2^{18}\}$ 
    \newline $T(N) = O(N)$ \\
    \midrule
    25 & Activate Function & Relu & The Relu operator is calculated by applying the Relu function to a sequence of length $N$:
    $\textit{Relu}(x) = \max(0, x)$ & $N = \{2^{10},2^{12},2^{14},2^{16},2^{18}\}$ 
    \newline $T(N) = O(N)$ \\
    \midrule
    26 & Activate Function & Leaky Relu & The Leaky Relu operator is calculated by applying the Leaky Relu function to a sequence of length $N$ with a small slope $\alpha$:
    $\textit{Leaky\_relu}(x) = x \text{ if } x > 0, \text{ otherwise } x \cdot \alpha$ & $N = \{2^{10},2^{12},2^{14},2^{16},2^{18}\}$ 
    \newline $T(N) = O(N)$ \\
    \midrule
    27 & Loss Function & MSE Loss & Given two vectors ${X}$ and ${Y}$, each of length $N$, calculate their mean squared error $\textit{MSE}({X}, {Y})$:
    $\textit{MSE}({X}, {Y}) = \frac{1}{N} \sum_{i=1}^{N} ({X}[i] - {Y}[i])^2$ & $N = \{2^{10},2^{12},2^{14},2^{16},2^{18}\}$ 
    \newline $T(N) = O(N)$ \\
    28 & Loss Function & Hinge Loss & The hinge loss operator is given a sequence of length $N$, and two input arrays: $\textit{predictions}$ and $\textit{targets}$. The $\textit{predictions}$ array contains the predicted values, and the $\textit{targets}$ array contains the true binary labels ($0$ or $1$). The operator calculates the hinge loss for each sample in the sequence using the formula:
    if $\textit{target} = 1$, $\textit{hinge\_loss}(x) = \max(0, 1 - \textit{pred})$
    if $\textit{target} = 0$, $\textit{hinge\_loss}(x) = \max(0, 1 + \textit{pred})$ & $N = \{2^{10},2^{12},2^{14},2^{16},2^{18}\}$ 
    \newline $T(N) = O(N)$ \\
    \midrule
    29 & Loss Function & Cross Entropy Loss & Given a predicted probability distribution ${X}$ of length $N$, where $N$ is the number of classes, and an actual probability distribution ${Y}$ of length $N$, compute their cross entropy loss $\textit{cross\_entropy\_loss}({X}, {Y})$:
    $\textit{cross\_entropy\_loss}({X}, {Y}) = -\sum_{i=1}^{N} {Y}[i] \cdot \log({X}[i])$ & $N = \{2^{14},2^{16},2^{18},2^{20},2^{22}\}$ 
    \newline $T(N) = O(N)$ \\
    \midrule
    30 & LLM Operator Sequence & Softmax Attention & Given the query matrix ${Q}$ of size $[\textit{q\_seq\_len}, \textit{dim\_qk}]$, key matrix ${K}$ of size $[\textit{kv\_seq\_len}, \textit{dim\_qk}]$, and value matrix ${V}$ of size $[\textit{kv\_seq\_len}, \textit{dim\_v}]$, calculate the softmax attention by the formula:
    $Y = \textit{softmax}(\frac{QK^T}{\sqrt{\textit{dim\_qk}}})V$
    where softmax of shape $[N, C]$ is defined as:
    For each row $i$ (where $i = 0$ to $N-1$), compute:
    $\textit{row}_i = {x}[i, :]$
    $M = \max(\textit{row}_i)$
    $\textit{softmax}(\textit{row}_i) = \frac{\exp(\textit{row}_i - M)}{\sum(\exp(\textit{row}_i - M))}$
        & 
    $[\textit{q\_seq\_len}, \textit{kv\_seq\_len}, \textit{dim\_qk},\newline \textit{dim\_v}]$ = 
    \{[128, 256, 512, 512],
    [256, 512, 1024, 1024],
    [512, 1024, 2048, 2048],
    [1024, 2048, 4096, 4096],
    [256, 1024, 2048, 1024],
    [512, 512, 1024, 2048],
    [1024, 1024, 4096, 2048],
    [128, 2048, 512, 4096]\} 
    \newline $T(q\_seq\_len,kv\_seq\_len,dim\_qk$\newline $,dim\_v) = O(q\_seq\_len\times kv\_seq\_len\times(dim\_qk + dim\_v))$ \\
    \midrule
   31 & LLM Operator Sequence & Triangle Attention & Given the query, key, value matrix of shape $[\textit{batch\_size}, \textit{head\_num}, \textit{seq\_length}, \textit{head\_dim}]$ and an upper triangular matrix attention mask of shape $[\textit{batch\_size}, \textit{seq\_length}, \textit{seq\_length}]$, calculate the triangle attention matrix of shape $[\textit{batch\_size}, \textit{head\_num}, \textit{seq\_length}, \textit{head\_dim}]$.
    for each $b$ in $\textit{batch\_size}$ and $h$ in $\textit{head\_num}$:
    ${Q}, {K}, {V} = \textit{query}[b, h, :, :], \textit{key}[b, h, :, :], \textit{value}[b, h, :, :]$ \newline
    ${M} = \textit{mask}[b, :, :]$
    \newline
    ${S} = \frac{{Q} {K}^T}{ \sqrt{\textit{head\_dim}}}$
    \newline
    ${S}_\textit{masked} = {S} \odot {M}$
    \newline
    ${P} = \textit{tril\_softmax}({S}_\textit{masked})$ 
    \newline
    ${O} = {P} {V}$
        & 
    [\textit{batch\_size}, \textit{head\_num}, \textit{seq\_length}, \textit{head\_dim}] = 
    \{[1, 32, 512, 128],
    [1, 96, 256, 128],
    [1, 48, 1024, 128],
    [2, 16, 256, 256],
    [2, 64, 512, 128],
    [4, 8, 32, 128], 
    [4, 32, 512, 128],
    [8, 16, 512, 128]\} 
    \newline $T(batch\_size, head\_num,$\newline $seq\_length,head\_dim) = O(batch\_size\times head\_num \times seq\_length^2 \times head\_dim)$ \\
    32 & LLM Operator Sequence & Merge Attention States & The merge\_attn\_states operator will accept two tensors ${V}_a$ and ${V}_b$ of the same shape $[\textit{num\_tokens}, \textit{num\_heads}, \textit{head\_size}]$, and two $\textit{LSE}_a$ and $\text{LSE}_b$ of the same shape $[\textit{num\_heads}, \textit{num\_tokens}]$, and perform merge exponential normalization:
    For each head $h$ and token $t$:
    $\textit{max\_lse} = \max(\textit{LSE}_a[h, t], \textit{LSE}_b[h, t])$
    $w_a = \exp(\textit{LSE}_a[h, t] - \textit{max\_lse})$
    $w_b = \exp(\textit{LSE}_b[h, t] - \textit{max\_lse})$
    $\textit{total\_w} = w_a + w_b$
    
    ${V}_\textit{out}[t, h, :] = ({V}_a[t, h, :] \cdot w_a + {V}_b[t, h, :] \cdot w_b) / \textit{total\_w}$
    $\textit{LSE}_\textit{out}[h, t] = \log(\textit{total\_w}) + \textit{max\_lse}$
        & 
    \textit{num\_tokens} = \{512, 768, 1024\} \newline
    \textit{num\_heads} = \{32, 64\}\newline
    \textit{head\_size} = \{128, 256\} 
    \newline $T(num\_tokens, num\_heads,$\newline $head\_size) = O(num\_tokens\times num\_heads \times head\_size)$ \\
    \midrule
    33 & LLM Operator Sequence & SwiGLU & The SwiGLU operator is calculated by applying the Swish-Gated Linear Unit function to two sequences $\text{gate}\ {g}$ and $\text{value}\ {v}$ of length $N$:
    \newline
    $\textit{Swish}(x) = x \cdot \sigma(\beta x) = \frac{x}{1 + e^{-\beta x}}$
    $\textit{SwiGLU}({g}, {v}) = \textit{Swish}({g}) \cdot {v}$ & $N = \{2^{10},2^{12},2^{14},2^{16},2^{18}\}$ 
    \newline $T(N) = O(N)$\\
    \midrule
   34 & LLM Operator Sequence & Silu and Mul & Given a concatenated input in shape $(B, 2D)$, where the first $D$ is ${x}$ and the second $D$ is ${g}$. Compute silu of ${x}$ and multiply it with ${g}$.
    ${x} = {input}[b, :D]$ \newline
    ${g} = {input}[b, D:]$
    ${out} = \textit{silu}({x}) \cdot {g}$
        & $B = \{16, 32, 64\}$ \newline
    $D = \{4096, 5120, 8192, 12288\}$
    \newline $T(B,D) = O(BD)$
     \\
    \midrule
    35 & LLM Operator Sequence & Rope & Given an input tensor ${x}$ of shape $(M, D)$, calculate the Rotary Positional Embedding. For each position $m$ where $m = 0$ to $M-1$, its $D$-dimensional vector ${v}_m = [x_{m,0}, x_{m,1}, ..., x_{m,D-1}]$. For each two-dimensional pair $(x_{m,2k}, x_{m,2k+1})$, where $k = 0$ to $(D/2)-1$:
    $\theta_k = \textit{base}^{-2k / D}$, $\phi(m,k) = m \cdot \theta_k$ \newline
    $\textit{out}[m,2k] = x_{m,2k} \cdot \cos(\phi(m,k)) - x_{m,2k+1} \cdot \sin(\phi(m,k))$ \newline
    $\textit{out}[m,2k+1] = x_{m,2k} \cdot \sin(\phi(m,k)) + x_{m,2k+1} \cdot \cos(\phi(m,k))$ & 
    $M = \{2^{4},2^{5},2^{6},2^{7}\}$
    \newline
    $D = \{2^{8},2^{10},2^{12},2^{14}\}$ 
    \newline $T(M,D) = O(MD)$\\
    \midrule
    36 & Tensor Process & Reverse Array & Reverse an input sequence of length $N$ into an output sequence of length $N$. & $N = \{2^{10},2^{12},2^{14},2^{16},2^{18}\}$ 
    \newline $T(N) = O(N)$ \\
    \midrule
    37 & Tensor Process & Matrix Copy & Given two matrices $A$ and $B$, copy $A$ into $B$, where matrix $A$ and $B$ has the same shape of $(M, N)$. & $M = \{2^8,2^{10},2^{12},2^{14},2^{16}\}$\newline $N = 2048$ 
    \newline $T(M,N) = O(MN)$\\
    \midrule
    38 & Tensor Process & Matrix Transpose & Transpose a matrix $A$ of dimension $(M,N)$ and output the transposed matrix $A^T$, where $A^T$ has dimensions $(N,M)$. & $M = \{2^{10},2^{11},2^{12},2^{13},2^{14}\}$\newline $N = \{2^{10},2^{11},2^{12},2^{13},2^{14}\}$ 
    \newline $T(M,N) = O(MN)$ \\
   39 & Tensor Process & High Order Contraction & Calculate the high order tensor contraction of tensor $A$ of shape $[a_{\textit{dim}}, x_{\textit{dim}}, b_{\textit{dim}}, y_{\textit{dim}}]$ and $B$ of shape $[x_{\textit{dim}}, c_{\textit{dim}}, y_{\textit{dim}}]$. Write the answer into tensor $C$ of shape $[a_{\textit{dim}}, b_{\textit{dim}}, c_{\textit{dim}}]$
    $C[a,b,c] = \sum_{x,y} A[a,x,b,y] \times B[x,c,y]$
    & $[a_{\textit{dim}}, b_{\textit{dim}}, c_{\textit{dim}}] $ = \{[32, 32, 32], [64, 32, 48], [128, 128, 64], [256, 256, 128]\} \newline $[x_{\textit{dim}}, y_{\textit{dim}}] $ = \{[4, 4],[16, 8], [16, 16]
    \}
    \newline $T(a_{dim},b_{dim},c_{dim},x_{dim},y_{dim}) = O(a_{dim}b_{dim}c_{dim}x_{dim}y_{dim})$ \\
     \midrule
    40 & Stencil Computation & 2D Max Pooling & Given an input signal of shape $[N, C, H, W]$. For each output position $(n, c, h_{\textit{out}}, w_{\textit{out}})$, compute the maximum value over the corresponding input window: \newline
    $\textit{output}[n, c, h_{\textit{out}}, w_{\textit{out}}]$ = $\max(\textit{input}[n, c, h:h+\textit{kernel\_size}, w:w+\textit{kernel\_size}])$
    \newline
    where $h = h_{\textit{out}} \times \textit{stride}$ and $w = w_{\textit{out}} \times \textit{stride}$ \newline
    The boundary values are filled with padding. & $[N, C, H, W]$ = \{[8, 3, 512, 512], [4, 3, 224, 224], [16, 3, 608, 608], [4, 1, 2048, 196]\} \newline [\textit{kernel\_size}, \textit{stride}] = \{[2,1], [3,1], [4,2]\}, \textit{padding} = \{0,1\} \newline
    $H_{out} = \lfloor\frac{(H + 2\times\textit{padding} - \textit{kernelsize})}{\textit{stride}} + 1\rfloor$
    $W_{out} = \lfloor\frac{(W + 2\times\textit{padding} - \textit{kernelsize})}{\textit{stride}} + 1\rfloor$
    \newline $T(N,C,H_{out},W_{out},kernel\_size) = O(NCH_{out}W_{out}\times kernel\_size^2)$ \\
    \midrule
    41 & Stencil Computation & 1D Convolution & Given an 1d input signal and an 1d kernel, compute the output signal using 1D convolution. & \textit{input\_size} = $\{2^{14},2^{16},2^{18},2^{20},2^{22}\}$ \newline 
    \textit{kernel\_size} = $\{32,64,128,256\}$
    \newline $T(input\_size,kernel\_size) = O(input\_size\times kernel\_size)$ \\
    \midrule
    42 & Stencil Computation & 2D Convolution & Given an 2d input signal and an 2d kernel, compute the output signal using 2D convolution. & \textit{input\_size} = $\{[2^8,2^8],[2^{10},2^{10}],[2^{12},2^{12}]\}$ \newline 
    \textit{kernel\_size} = $\{[8,8],[16,16],[32,32]\}$
    \newline $T(input\_size\_x,input\_size\_y,$\newline$kernel\_size\_x,kernel\_size\_y) = O(input\_size\_x\times input\_size\_y \times kernel\_size\_x \times kernel\_size\_y)$ \\
    \midrule
    43 & Stencil Computation & 2D Stencil & Given an input $u_\textit{old}$ of shape $[n_x, n_y]$,calculate the 2d 5-point stencil of $u_\textit{old}$ and store the result in $u_\textit{new}$:
    For interior points ($i \in [1, n_x-2]$, $j \in [1, n_y-2]$):
    $u_{\textit{new}}[i, j] = u_{\textit{old}}[i, j] + r \ (u_{\textit{old}}[i-1, j] + u_{\textit{old}}[i+1, j] + u_{\textit{old}}[i, j-1] + u_{\textit{old}}[i, j+1] - 4 \ u_{\textit{old}}[i, j])$
    For boundary points $(i \in \{0, n_x-1\}, j \in \{0, n_y-1\})$:
    $u_\textit{new}[i, j] = u_\textit{old}[i, j]$ & $n_x = \{2^{10},2^{12},2^{14}\}$ \newline $n_y = \{2^{10},2^{12},2^{14}\}$
    \newline $T(n_x,n_y) = O(n_xn_y)$ \\
   44 & Stencil Computation & 3D Stencil & Given an input $u_{\textit{old}}$ of shape $[n_x, n_y, n_z]$, calculate the 3D 7-point stencil of $u_{\textit{old}}$ and store the result in $u_{\textit{new}}$:
    For interior points ($i \in [1, n_x-2]$, $j \in [1, n_y-2]$, $k \in [1, n_z-2]$):
    $u_{\textit{new}}[i, j, k] = u_{\textit{old}}[i, j, k] + r \times (
    u_{\textit{old}}[i-1, j, k] + u_{\textit{old}}[i+1, j, k] +
    u_{\textit{old}}[i, j-1, k] + u_{\textit{old}}[i, j+1, k] +
    u_{\textit{old}}[i, j, k-1] + u_{\textit{old}}[i, j, k+1] -
    6.0 \times u_{\textit{old}}[i, j, k])$
    For boundary points (where $i \in {0, n_x-1}$ OR $j \in {0, n_y-1}$ OR $k \in {0, n_z-1}$):
    $u_{\textit{new}}[i, j, k] = u_{\textit{old}}[i, j, k]$ & $n_x = \{2^{6},2^{7},2^{8}\}$ \newline $n_y = \{2^{6},2^{7},2^{8}\}$ \newline $n_z = \{2^{6},2^{7},2^{8}\}$
    \newline $T(n_x,n_y,n_z) = O(n_xn_yn_z)$ \\
    \midrule
    45 & Math \& Algorithm & Sorting & Sort an input array of length $N$ in ascending order. The naive implementation is merge sort. You are free to choose any sorting algorithm.
    & $N = \{2^{8},2^{10},2^{12},2^{14},2^{16}\}$ 
    \newline $T(N) = O(N\log N)$ \\
    \midrule
   46 & Math \& Algorithm & Top K & Given a vector ${X}$ of length $N$, calculate the top $k$ of ${X}$ and their indices, which means finding the $k$ largest elements in the sequence and their indices. The naive implementation is merge sort. You are free to choose any sorting algorithm ensuring values in descending order and index in ascending order.
    & $N = \{2^{8},2^{10},2^{12},2^{14},2^{16}\}$ \newline 
    $k = \{32, 64, 128\}$
    \newline $T(N) = O(N\log N)$ \\
    \midrule
    47 & Math \& Algorithm & Simpson Int & Calculate the definite integral using the Simpson method, with $N$ sample points $y_\textit{samples}$ and upper and lower limits of integration $b$ and $a$ respectively:

    $\int_a^b f(x)dx \approx \frac{h}{3} [ y_0 + y_{N-1} + $ $4\sum_{i=1,3,5,\ldots}^{N-2} y_i + 2\sum_{i=2,4,6,\ldots}^{N-3} y_i ]$
    & 
    $N = \{2^{14},2^{16},2^{18},2^{20},2^{22}\}$ 
    \newline $T(N) = O(N)$\\
    \midrule
    48 & Math \& Algorithm & Monte Carlo Int & Calculate the definite integral using the Monte Carlo method, with $N$ sample points $y_\textit{samples}$ and upper and lower limits of integration $b$ and $a$ respectively:

   $\int_a^b f(x)dx \approx \frac{b - a}{N}\sum_{i=1}^N y_i$
    & 
    $N = \{2^{14},2^{16},2^{18},2^{20},2^{22}\}$ 
    \newline $T(N) = O(N)$\\
    \midrule

   49 & Math \& Algorithm & FFT & Given a complex number array of length $N$, calculate the FFT of the array. The input and output are separated into real and imaginary parts.
    & $N = \{2^{10},2^{11},2^{12},2^{13},2^{14}\}$
    \newline $T(N) = O(N\log N)$ \\
    \midrule
    50 & Math \& Algorithm & Prefix Sum & Given a sequence of length $N$, calculate the prefix sum of the sequence on each index and write into an output sequence. The naive implementation computes each output element independently by summing all elements from the beginning of the input sequence up to the current index.
    & $N = \{2^{10},2^{12},2^{14},2^{16},2^{18}\}$
    \newline $T(N) = O(N^2)$  \\
    \end{longtable}

\end{document}